\documentclass[conference]{IEEEtran}
\usepackage{times}

\usepackage[numbers,sort&compress]{natbib}
\usepackage{multicol}
\usepackage[bookmarks=true]{hyperref}

\usepackage{url}
\usepackage{xcolor,colortbl}
\usepackage{amsmath,amsfonts,amssymb,wasysym}
\usepackage{verbatim}
\usepackage{mathrsfs}
\let\chapter\section
\usepackage{wrapfig}
\usepackage{graphicx} 
\usepackage{xspace}
\usepackage{capt-of}
\usepackage[ruled,linesnumbered]{algorithm2e}

\usepackage{multirow,multicol}
\usepackage{color}
\usepackage{soul}
\usepackage{hyperref}

\SetKwComment{Comment}{$\triangleright$\ }{}

\DeclareMathOperator{\diag}{diag}

\begin{document}
\title{Learning to Slide Unknown Objects \\with Differentiable Physics Simulations}

\author{\authorblockN{Changkyu Song and Abdeslam Boularias}
\authorblockA{Computer Science Department, Rutgers University, New Jersey, USA}
\authorblockA{\{cs1080,abdeslam.boularias\}@cs.rutgers.edu}
}

\maketitle

\begin{abstract}
 We propose a new technique for pushing an unknown object from an initial configuration to a goal configuration with stability constraints. The proposed method leverages recent progress in differentiable physics models to learn unknown mechanical properties of pushed objects, such as their distributions of mass and coefficients of friction. 
 The proposed learning technique computes the gradient of the distance between predicted poses of objects and their actual observed poses, and utilizes that gradient to search for values of the mechanical properties that reduce the reality gap. The proposed  approach is also utilized to optimize a policy to efficiently push an object toward the desired goal configuration.
 Experiments with real objects using a real robot to gather data show that the proposed approach can identify mechanical properties of heterogeneous objects from a small number of pushing actions.
\end{abstract}

\section{Introduction}
Nonprehensile manipulation of objects is a practical skill used frequently by humans to displace objects from an initial configuration to a desired final one with a minimum effort. In robotics, this type of manipulation can be more advantageous than the traditional pick-and-place when an object cannot be easily grasped by the robot, due to the design of the end-effector and the size of the object, or the obstacles surrounding the manipulated object. For example, combined pushing and grasping actions have been shown to succeed where traditional grasp planners fail, and to work well under uncertainty~\cite{Dogar2011AFF,Dogar2012,king2015,king2016,KingICRA2017,Pinto-abs-1810-10654,DBLP:conf/icra/YuanSKWH18}.

 The mechanics of planar pushing was extensively explored in the past~\cite{Mason86}. Large datasets of images of planar objects pushed by a robot on a flat surface were also recently presented~\cite{fazeli2017ijrr,bauza2019iros}. Recent techniques for planar sliding mechanics focus on learning data-driven models for predicting the motions of the pushed objects in simulation. While this problem can be solved to a certain extent by using generic end-to-end machine learning tools such as neural networks~\cite{bauza2019iros}, model identification methods that are explicitly derived from the equations of motion are generally more efficient~\cite{JJZhou2018}. 
 A promising new direction is to directly differentiate the prediction error with respect to the model of the object's mechanical properties, such as its mass and friction distributions, and to use standard gradient descent algorithms 
to search for values of the properties that reduce the gap between simulated motions and observed ones. 
Unfortunately, most popular physics engines do not natively provide the derivatives of the predicted poses~\cite{ErezTT15,DART,PhysX,Bullet,ODE}, and 
the only way to differentiate them is through numerical finite differences, which are expensive computationally. 

 \begin{figure}[t]
    \centering
        \includegraphics[width=1\linewidth]{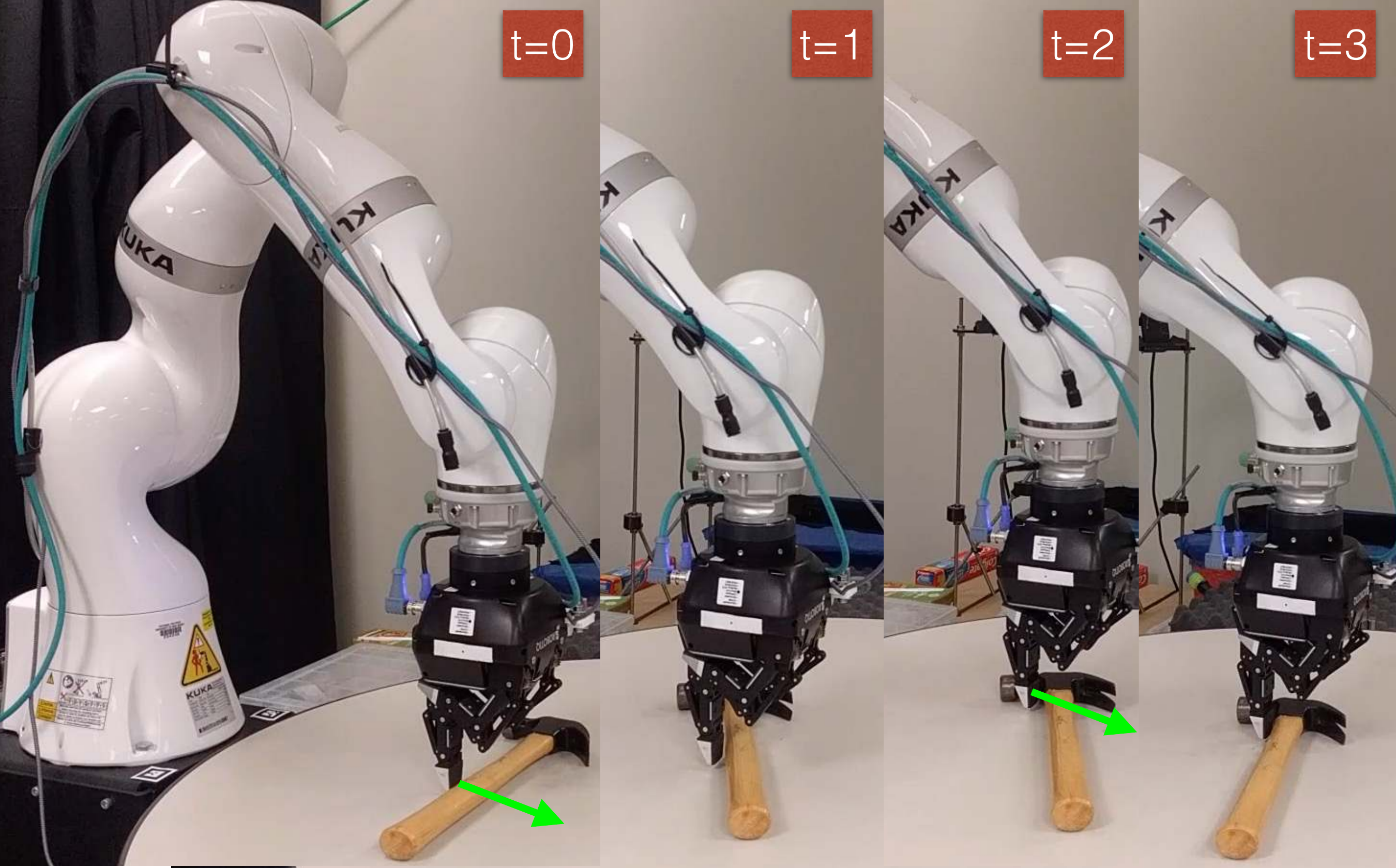} 
    \caption{Robotic setup used for pre-grasp sliding experiments. In this example, the robot pushes a hammer into a desired final pose on the edge of the table through consecutive pushes. The choice of contact points, pushing directions and a stable goal configuration is non-trivial as it depends on the mass distribution and the friction coefficients of different parts of the object, which are unknown {\it a priori}. This work proposes a method for identifying these parameters from a small number of observed motions of the object.}
    \label{fig:robot}
    \vspace{-0.7cm}
\end{figure}

In the present work\footnote{This work was supported by NSF awards 1734492, 1723869 and 1846043.}
, we propose a method for safely sliding an unknown object from an initial configuration to a target one. The proposed approach integrates a model identification algorithm with a planner. 
To account for non-uniform surface properties and mass distributions, the object is modeled as a large set of small cuboids that may have different material properties and that are attached to each other with fixed rigid joints. A simulation error function is given as the distance between the centers of the objects in simulated trajectories and the true observed ones. The gradient of the simulation error is used to search for the object's mass and friction distributions. 

The {\it  main contribution} is the derivation of the analytical gradient of the simulation error with respect to the mass and friction distributions using the proposed cuboid representation of objects.
The {\it  second contribution} is the use of the derived analytical gradients for identifying models of unknown objects by using a robotic manipulator, and demonstrating the computational and data efficiency of the proposed approach. 
The {\it  third contribution} is the use of the proposed integrated method for pre-grasp sliding manipulation of thin unknown objects.

A video of the data collection process, the pre-grasp sliding and grasping experiments using the robotic setup shown in Figure~\ref{fig:robot} can be viewed at {\textcolor{blue}{\url{https://bit.ly/371Y6Y1}}} .

\section{Related Work}
Algorithms for \emph{Model-based reinforcement learning} explicitly learn the unknown dynamics, often from scratch, and search for an optimal policy accordingly~\cite{Dogar_2012_7076,LunchMason1996,isbell:physics:2014,ZhouPBM16,abbeel2006using,DBLP:conf/aaai/BoulariasBS15}.
The unknown dynamics are often modeled using an off-the-shelf statistical learning algorithm, such as a {\it Gaussian Process} (GP)~\cite{bauza*2018corl,Deisenroth:2011fu,Calandra2016,MarcoBHS0ST17,bansal2017goal,pautrat2017bayesian,8624443}, or a neural network~\cite{OhGLLS15,chiappa2017recurrent,finn2016deep,Finn2016DeepSA}. This approach was recently used to collect images of pushed objects and build models of their motions~\cite{yu2016iros,bauza*2018corl,doshi2020icra,bauza2019iros}. While the proposed method belongs to the category of model-based RL, it differs from most related methods by the explicit use of the dynamics equations, which drastically improves its data-efficiency. The identified mass distribution can also be used to predict the balance and stability of the object in new configurations that are not covered in the training data. For instance, a GP or a neural net cannot predict if an object remains stable when pushed to the edge of a support table, unless such an example is included in the training data, with the risk dropping the object and losing it.

The {\it mechanics of pushing} was explored in several past works~\cite{Mason86,doi:10.1177/027836499601500602,23847993b652419a91558fd1f03bbec3,doi:10.1177/027836499601500603,Dogar2010PushgraspingWD,DBLP:conf/icra/ZhouPBM16,DBLP:journals/ijrr/ZhouHM19,DBLP:journals/ijrr/ZhouMPB18,Yoshikawa1991IndentificationOT,ShaojunIJCAI2018,L4DC2020Changkyu}, from both a theoretical and algorithmic point of view. 
Notably, Mason~\cite{Mason86} derived the voting theorem to predict the rotation and translation of an object pushed by a point contact. A strategy for stable pushing when objects remain in contact with an end-effector was also proposed in~\cite{doi:10.1177/027836499601500602}. Yoshikawa and Kurisu~\cite{Yoshikawa1991IndentificationOT} proposed a regression method for identifying the support points of a pushed object by dividing the support surface into a grid and approximating the measured frictional force and torque as the sum of unknown  frictional forces applied on the grid's cells. A similar setup was considered in~\cite{23847993b652419a91558fd1f03bbec3} with a constraint to ensure positive friction coefficients. The {\it limit surface} is a convex set of all friction forces and torques that can be applied on an object in quasi-static pushing. The limit surface is often approximated as an ellipsoid~\cite{doi:10.1177/027836499601500603}, or a higher-order convex polynomial~\cite{DBLP:conf/icra/ZhouPBM16,DBLP:journals/ijrr/ZhouHM19,DBLP:journals/ijrr/ZhouMPB18}. An ellipsoid approximation was also used to simulate the motion of a pushed object to perform a {\it push-grasp}~\cite{Dogar2010PushgraspingWD}.
In contrast with our method, these works identify only the friction parameters, and assume that the mass distribution is known or irrelevant in a quasi-static regime. 

There has been a recent surge of interest in developing \emph{natively differentiable physics engines}~\cite{DegraveHDW16,DBLP:journals/corr/Al-RfouAAa16}. 
A combination of a learned and a differentiable simulator was used to predict effects of actions on planar objects~\cite{DBLP:journals/corr/abs-1710-04102}, and to learn fluid parameters~\cite{Schenck2018SPNetsDF}.
Differentiable physics simulations were also used for manipulation planning and tool use~\cite{18-toussaint-RSS}. 
Recently, it has been observed that a standard physical simulation, formulated as a Linear Complementary Problem (LCP), is also differentiable and can be implemented in PyTorch~\cite{Belbute-Peres2017}. In~\cite{Mordatch:2012}, a differentiable contact model was used to allow for optimization of several  locomotion and manipulation tasks.

\section{Problem Setup and Notation}
\label{problem_setup}
\begin{figure*}
{
\begin{equation*}
\boxed{
\begin{bmatrix} \mathbf{0} \\ \mathbf{0}  \\ \rho \\ \xi \end{bmatrix}
+ \begin{bmatrix} \textcolor{red}{\mathcal{M}} & \mathcal{J}^T_e(x_t) & \mathcal{J}^T_f(\dot{x}_t) & \mathbf{0} \\
\mathcal{J}_e(x_t) & \mathbf{0} & \mathbf{0} & \mathbf{0}\\ 
\mathcal{J}_f(\dot{x}_t) & \mathbf{0} & \mathbf{0} & -I \\ 
\mathbf{0} & \mathbf{0} & I & \mathbf{0} \end{bmatrix}
\begin{bmatrix} -\dot{x}_{t+dt} \\ \lambda_e \\ \lambda_f \\ \gamma \end{bmatrix}
=
\begin{bmatrix} -\textcolor{red}{\mathcal{M}}\dot{x}_{t}-dt \textcolor{blue}{{F_t}} \\ \mathbf{0}  \\ \mathbf{0} \\ \textcolor{red}{{\mu}}\diag(\textcolor{red}{\mathcal{M}}) \end{bmatrix}
\textrm{s.t.}
\begin{bmatrix} \rho \\ \xi  \end{bmatrix} \geq \mathbf{0},
\begin{bmatrix} \lambda_f \\ \gamma  \end{bmatrix} \geq \mathbf{0},
\begin{bmatrix} \rho \\ \xi  \end{bmatrix} \begin{bmatrix} \lambda_f \\ \gamma  \end{bmatrix} = \mathbf{0}
}
\end{equation*}
}
\vspace{-0.3cm}
\caption{Equations of Motion as a Linear Complementarity Problem (LCP)}
\label{equation_motion}
\end{figure*}

We consider the problem of displacing a rigid object on a flat homogeneous surface from an initial pose $x_0$ to a desired final pose $x^d_T$. 
The object has an unknown shape and material properties. We assume that a depth-sensing camera provides a partial 3D view of the object. The partial view contains only the object's upper surface. 
A 3D shape is then automatically constructed  by assuming that the occluded bottom side is flat. 
Most of the objects used in our experiments are not flat. However, the autonomously learned friction model  simply assigns near-zero friction coefficients to the regions of the bottom surface that do not actually touch the tabletop. Thus, learned near-zero friction forces compensate for wrongly presumed flat regions in the occluded bottom part of the object. 

We approximate the object as a finite set of small cuboids. The object is divided into large number of connected cells $1,2,\dots,n$, using a regular grid structure. Each cell $i$ has its own local mass and coefficient of friction that can be different from the other cells. 
The object's pose $x_t$ at time $t\in [0,T]$ is a vector in $[SE(2)]^n$ corresponding to the translation and rotation in the plane for each of the $n$ cells. In other terms, $x_t = [p^1_{x,t},p^1_{y,t},\theta^1_t,\dots, p^n_{x,t},p^n_{y,t},\theta^n_t]^T$, where $(p^i_{x,t},p^i_{y,t})$ is the $i^{th}$ cell's 2D position on the surface, and $\theta^i_t$ is its angle of rotation. Similarly, we denote the object's generalized velocity (a {\it twist}) at time $t$ by $\dot{x}_t = [\dot{p}^i_{x,t},\dot{p}^i_{y,t},\dot{\theta}^i_t]_{i=0}^n$, where $(\dot{p}^i_{x,t},\dot{p}^i_{y,t})$ is the $i^{th}$ cell's linear velocity on the surface, and $\dot{\theta}^i_t$ is its angular velocity. 
The object's mass matrix $\mathcal{M}$ is a diagonal $3n\times3n$ matrix, where the diagonal is $[\mathcal{I}_1,\mathcal{M}_1,\mathcal{M}_1,\mathcal{I}_2,\mathcal{M}_2,\mathcal{M}_2,\dots, \mathcal{I}_n,\mathcal{M}_n,\mathcal{M}_n]$, $\mathcal{I}_i$ is the moment of inertia of the $i^{th}$ cell of the object, and $\mathcal{M}_i$ is its mass. $\mathcal{I}_i = \frac{1}{6} \mathcal{M}_i w^2$ where $w$ is the width of a cuboid.
$\mu$ is a $3n\times 3n$ diagonal matrix, where the diagonal is $[\mu_1,\mu_1,\mu_1,\mu_2,\mu_2,\mu_2,\dots,\mu_n,\mu_n,\mu_n]$. $\mu_i$ is the coefficient of friction between the $i^{th}$ cell of the object and the support surface. We assume that: $\forall i\in \{1,\dots, n\}: \mu_i\in[0,\mu_{max}]$, where $\mu_{max}$ is a given upper bound. 
An external generalized force ({\it a wrench}) denoted by $F$ is an $1\times 3n$ vector $[f^1_{x},f^1_{y},\tau^1,\dots, f^n_{x},f^n_{y},\tau^n]^T$, where $[f^i_{x},f^i_{y}]$ and $\tau^n$ are respectively the force and torque applied on cell $i$. External forces are generated from the contact between the object and a fingertip of the robotic hand used to push the object. We assume that at any given time $t$, at most one cell of the object is in contact with the fingertip. Therefore, $F = [0,0,0,\dots,f^{c(t)}_{x},f^{c(t)}_{y},\tau^{c(t)},\dots, 0,0,0]^T$ where $c(t)\in\{0,\dots,n\}$ is the index of the contacted cell at time $t$.

A ground-truth trajectory $\mathcal T^g$ is a state-action sequence $(x^g_0,\dot{x}^g_0,F_{0}, \dots, x^g_{T-1},\dot{x}^g_{T-1}, F_{T-1}, x^g_{T},\dot{x}^g_{T})$, wherein $(x^g_t,\dot{x}^g_{t})$ is the observed pose and velocity of the pushed object, and $F_{t}$ is the external force applied at time $t$, as defined above.
A corresponding simulated trajectory $\mathcal T$ is obtained by starting at the same initial state
$\hat{x}_0$ in the
corresponding real trajectory, i.e., $\hat{x}_0 = x_0$, and applying
the same control sequence $(F_0, F_1, \dots, F_{T-1})$. Thus,
the simulated trajectory $\mathcal T$ results in a state-action sequence $(x^g_0,\dot{x}^g_0,F_{0},x_1,\dot{x}_1,F_{1}, \dots, x_{T-1},\dot{x}_{T-1}, F_{T-1}, x_{T},\dot{x}_{T})$,
where $x_{t+1} = x_{t} + \dot{x}_{t}dt$ is the predicted next pose. 
Velocity $\dot{x}_{t}$ is a vector corresponding to translation  and angular velocities in the plane for each of the $n$ cells, it is predicted in simulation as
$\dot{x}_{t+1} = V(x_{t},\dot{x}_{t}, F_t, \mathcal M, \mu)$.
The goal is to identify mass distribution $\mathcal M$ and friction map $\mu$ that result in simulated trajectories that are as close as possible to the real observed ones.  Therefore, the objective is to solve the following optimization problem,
{\small
\begin{eqnarray}
(\mathcal M^*, \mu^*) &=& \arg \min_{\mathcal M, \mu} loss(\mathcal M, \mu),  \label{simulationError} \\ loss(\mathcal M, \mu)\hspace{-0.2cm} &\stackrel{def}{=}& \hspace{-0.2cm}  \sum_{t=0}^{T-2}\|x^g_{t+2} - \big(x^g_{t+1} +  V(x^g_{t},\dot{x}^g_{t}, F_t, \mathcal M, \mu)dt \big) \|_2. \nonumber
\end{eqnarray}
}
Since $x_{t}$ is a vector containing all cells' positions, the loss is the sum of distances between each cell's ground-truth pose and its predicted pose, which is equivalent to the average distance (ADD) metric as proposed in~\cite{Hinterstoisser2013}.
In the following, we explain how velocity function $V$ is computed.


\section{Forward Simulation}
We adopt here the formulation presented in~\cite{cline,Belbute-Peres:2018:EDP:3327757.3327820,Belbute-Peres2017}. 
We adapt and customize the formulation to exploit the proposed grid-structure representation, and we extend it to include frictional forces between a pushed object and a support surface. 
The transition function is given as $\dot{x}_{t+1} = x_t + \dot{x}_t dt$ where $dt$ is the duration of a constant short time-step. Velocity $\dot{x}_t$ is a function of force $F_t$ and mechanical parameters $\mathcal M$ and $\mu$. 
To find $\dot{x}_{t+1}$, we solve the system of equations of motion that we present in Figure~\ref{equation_motion}, where $x_t$ and $\dot{x}_{t}$ are inputs, $[\rho,\xi]$ are slack variables, $[\dot{x}_{t+dt},\lambda_e,\lambda_f,\gamma]$ are unknown vectors, and
$[\mathcal{M}, \mu]$ are hypothesized mass and friction matrices.  $\diag(\mathcal{M})$ is a $1\times 3n$ vector corresponding to the main diagonal of $\mathcal{M}$. 

$\mathcal{J}_e (x_t)$ is a global Jacobian matrix of all the adjacency constraints in the grid structure. These constraints ensure that the different cells of the object move together with the same velocity. $\mathcal{J}_e (x_t)$ is an $m\times n$ matrix where $n$ is the number of cells, and $m$ is the number of pairs of adjacent cells. 
\begin{eqnarray*}
\small 
\mathcal{J}_e (x_t) = 
\begin{bmatrix} 
\mathcal{J}^{1,1}_e (x_t) & \dots & \mathcal{J}^{1,n}_e (x_t) \\
\vdots & \dots & \vdots \\
\mathcal{J}^{m,1}_e (x_t) & \dots & \mathcal{J}^{m,n}_e (x_t) \end{bmatrix} , 
\end{eqnarray*}
If cell $i$, whose four sides have length $l$, is one of the two adjacent cells in the pair indexed by $k$, then
\begin{eqnarray*}
\small 
\mathcal{J}^{k,i}_e (x_t) = 
\begin{bmatrix} 
- \frac{l}{2}\sin{\theta^i_t} & 1 & 0 \\
  \frac{l}{2}\cos{\theta^i_t} & 0 & 1 
\end{bmatrix}. \textrm{Else}, 
\mathcal{J}^{k,i}_e (x_t) = 
\begin{bmatrix} 
0 & 0 & 0 \\
0 & 0 & 0 
\end{bmatrix}. 
\end{eqnarray*}
$\lambda_e$ is a $2m\times 1$ variable vector that is multiplied by the Jacobian $\mathcal{J}_e(x_t)$ to generate the vector of impulses $\mathcal{J}^T_e(x_t)\lambda_e$, which are time-integrals of internal forces that preserve the rigid structure of the object. 

$\mathcal{J}_f(\dot{x}_t)$ is an $n\times n$ Jacobian matrix related to the frictional forces between the object's cells and the support surface, and the corresponding constraints. The main block-diagonal of 
$\mathcal{J}_f(\dot{x}_t)$ is $[\mathcal{J}^1_f(\dot{x}_t), \mathcal{J}^2_f(\dot{x}_t), \dots, \mathcal{J}^n_f(\dot{x}_t)]$, wherein
\begin{eqnarray*}
\small 
\mathcal{J}^i_f(\dot{x}_t) = 
\begin{bmatrix} 
-\textrm{sign}(\dot{\theta}^i_t) & 0 & 0\\
0 & \frac{\dot{p}^i_{x,t}}{\sqrt{(\dot{p}^i_{x,t})^2 + (\dot{p}^i_{y,t})^2}} & \frac{\dot{p}^i_{y,t}}{\sqrt{(\dot{p}^i_{x,t})^2 + (\dot{p}^i_{y,t})^2}} 
\end{bmatrix},
\end{eqnarray*}
and the remaining entries of $\mathcal{J}_f(\dot{x}_t)$ are all zeros. $\lambda_f$ is a $2n\times 1$ variable vector. $\mathcal{J}_f(\dot{x}_t)$ is multiplied by $\lambda_f$ to generate a vector of the frictional forces and torques between the support surface and each cell of the object. $\mathcal{J}_f(\dot{x}_t)$ defines the direction of the frictional forces and torques as the opposite of its current velocity $\dot{x}_t = [\dot{\theta}^i_t,\dot{p}^i_{x,t},\dot{p}^i_{y,t}]_{i=1}^n$, whereas $\lambda_f$ defines the scalar magnitudes of the frictional forces and torques.

The friction terms have complementary constraints, stated in Fig.~\ref{equation_motion}. These constraints are used to distinguish between the cases when the object is moving and friction magnitudes $\lambda_f$ are equal to $\mu\diag(\mathcal M)$, and the case when the object is stationary and the friction magnitudes $\lambda_f$ are smaller than $\mu\diag(\mathcal M)$. When the object moves, and assuming that the change in the direction of motion happens smoothly, we have $\mathcal{J}_f(\dot{x}_t) \dot{x}_{t+dt} < 0$. Therefore, $\gamma > 0$ because of the constraints $\rho =  \mathcal{J}_f(\dot{x}_t) \dot{x}_{t+dt} + \gamma I$ and $\rho \geq 0$ and $\gamma \geq 0$. Then $\xi = 0$ because of the constraint $\gamma\xi = 0$. We conclude that $\lambda_f = \mu\diag(\mathcal M)$ from the constraint $\xi + \lambda_f = \mu\diag(\mathcal M)$. Similarly, one can show that  $\lambda_f < \mu\diag(\mathcal M)$ if $\dot{x}_{t+dt} = 0$.

To simulate a trajectory $(x_0,\dot{x}_0,F_{0},x_1,\dot{x}_1,F_{1}, \dots)$, we iteratively find velocities $\dot{x}_{t+dt}$ by solving the equations in Fig.~\ref{equation_motion} where $({x}_{t},\dot{x}_{t}, F_t, \mu,\mathcal M)$ are fixed inputs and the remaining variables are unknown. 
The solution is obtained, after an initialization step, by iteratively minimizing the residuals from the equations in  Fig.~\ref{equation_motion}, using the convex optimizer of~\cite{MattingleySB12}.

\section{Mass and Friction Gradients}

To obtain material parameters $[\mathcal{M},\mu]$, a gradient descent on the loss function in Equation~\ref{simulationError} is performed. A first approach to compute the gradient is to use the {\it Autograd} library for automatic derivation in Python. We propose here a second simpler and faster approach based on deriving analytically the closed forms of the gradients $\frac{\partial loss (\mathcal{M},\mu)}{\partial \mu}$ and $\frac{\partial loss (\mathcal{M},\mu)}{\partial \mathcal{M}}$.

Let us denote by $(\dot{x}^*_{t+1},\lambda^*_e,\lambda^*_f,\gamma^*)$ the solutions for $(\dot{x}_{t+1},\lambda_e,\lambda_f,\gamma)$ in the system in Fig.~\ref{equation_motion}. 
Let us also use $D(x)$ to denote a matrix that contains a vector $x$ as a main diagonal and zeros elsewhere. Finally, let $\diag(\mathcal M)$ refer to the main diagonal of $\mathcal M$. In other terms,
$\diag(\mathcal M) = [\mathcal{I}_1,\mathcal{M}_1,\mathcal{M}_1,\dots, \mathcal{I}_n,\mathcal{M}_n,\mathcal{M}_n]$.
Then,
\begin{eqnarray*}
\footnotesize
\begin{cases}
-\mathcal{M}\dot{x}^*_{t+1}+\mathcal{J}^T_e(x_t) \lambda^*_e+\mathcal{J}^T_f(\dot{x}_t) \lambda^*_f+\mathcal{M}\dot{x}_{t}+F_t dt = 0,\\
-\mathcal{J}_e(x_t) \dot{x}^*_{t+1} = 0, \\
D(\lambda^*_f) (-\mathcal{J}_f(\dot{x}_t)  \dot{x}^*_{t+1}-\gamma^*) = 0,\\
D(\gamma^*) (\lambda^*_f- \mu\diag(\mathcal M)) = 0.
\end{cases}
\end{eqnarray*}
The differentials of the system are given as
\begin{eqnarray*}
\footnotesize
\hspace{-0.1cm}
\begin{cases}
-\partial\mathcal{M}\dot{x}^*_{t+1}-\mathcal{M}\partial\dot{x}_{t+1}+\mathcal{J}^T_e(x_t) \partial\lambda_e+\mathcal{J}^T_f(\dot{x}_t) \partial\lambda_f+\partial\mathcal{M}\dot{x}_{t} = 0,\\
-\mathcal{J}_e(x_t) \partial\dot{x}_{t+1} = 0, \\
D(-\mathcal{J}_f(\dot{x}_t)  \dot{x}^*_{t+1}-\gamma^*) \partial \lambda_f + D(\lambda^*_f) (-\mathcal{J}_f(\dot{x}_t)  \partial\dot{x}_{t+1}-\partial\gamma) = 0,\\
D (\lambda^*_f\hspace{-0.1cm}-\hspace{-0.1cm} \mu\diag(\mathcal M)) \partial\gamma \hspace{-0.1cm}+\hspace{-0.1cm}  D(\gamma^*) (\partial\lambda_f \hspace{-0.1cm}-\hspace{-0.1cm}  \partial\mu\diag(\mathcal M) \hspace{-0.1cm}-\hspace{-0.1cm}  \mu \partial\diag(\mathcal M)) = 0,
\end{cases}
\end{eqnarray*}
wherein $\partial x_t$, $\partial \dot{x}_t$, $\partial \mathcal{J}^T_e(x_t)$, $\partial \mathcal{J}^T_f(\dot{x}_t)$ are all zero matrices and vectors because ${x}_t$ and $\dot{x}_t$ are fixed and treated as a constant since they are set to ${x}^g_t$ and $\dot{x}^g_t$ in Equation~\ref{simulationError}. Also, $\partial F_t = \mathbf{0}$ because the applied force at time $t$ is given as a constant in the identification phase. 
The differentials can be arranged in the following matrix form: $G X = Y$,
\begin{eqnarray*}
\footnotesize
\hspace{-0.2cm}
\setlength{\arraycolsep}{0pt}
G \hspace{-0.1cm}=\hspace{-0.1cm}  
\begin{bmatrix} 
\mathcal{M} & \mathcal{J}^T_e(x_t) & \mathcal{J}^T_f(\dot{x}_t) & \mathbf{0} \\
\mathcal{J}_e(x_t) & \mathbf{0} & \mathbf{0} & \mathbf{0} \\
D(\lambda_f^*) \mathcal{J}_f(\dot{x}_t) & \mathbf{0} & D(-\mathcal{J}_f(\dot{x}_t) \dot{x}^*_{t+1} -\gamma^* ) & -D(\lambda_f^*)\\
\mathbf{0} & \mathbf{0} & D(\gamma^* ) & D(\lambda_f^* -\mu\diag(\mathcal M))\\
\end{bmatrix}
\end{eqnarray*}
\begin{eqnarray*}
X &=& [-\partial \dot{x}_{t+1} , \partial \lambda_e,\partial \lambda_f, \partial \gamma]^T, \\
Y &=& [\partial  \mathcal M (\dot{x}^*_{t+1} - \dot{x}_{t}), \mathbf{0} , \mathbf{0}, D(\gamma^*)\partial (\mu\diag(\mathcal M))]^T.
\end{eqnarray*}
Also,
\begin{eqnarray*}
&&\frac{\partial loss}{\partial (-\dot{x}_{t+1})} (-\partial \dot{x}_{t+1}) = [\frac{\partial loss}{\partial (-\dot{x}_{t+1})} , \mathbf{0} ,\mathbf{0} ,\mathbf{0} ]X \\
&&= [\frac{\partial loss}{\partial (-\dot{x}_{t+1})} , \mathbf{0} ,\mathbf{0} ,\mathbf{0} ]G^{-1} Y = [\alpha_1,\alpha_2,\alpha_3,\alpha_4] Y,
\end{eqnarray*}
wherein $[\alpha_1,\alpha_2,\alpha_3,\alpha_4]$ is defined as $[\frac{\partial loss}{\partial (-\dot{x}_{t+1})} , \mathbf{0} ,\mathbf{0} ,\mathbf{0} ]G^{-1}$. 
We use the blockwise matrix inversion to compute $G^{-1}$,
\begin{eqnarray*}
\footnotesize
\setlength{\arraycolsep}{5pt}
\begin{bmatrix} 
A & B \\
C & D \\
\end{bmatrix}^{-1}
=
\begin{bmatrix} 
(A-BD^{-1}C)^{-1} & -(A-BD^{-1}C)^{-1}BD^{-1} \\
g(A,B,C,D) & h(A,B,C,D)
\end{bmatrix}
\end{eqnarray*}
where $A$ and $D$ are square matrices, and $D$ and $(A-BD^{-1}C)$ are invertible.
Notice that to compute $[\alpha_1,\alpha_2,\alpha_3,\alpha_4]$, we only need the upper quarter of $G^{-1}$, because the remaining raws will be multiplied by $\mathbf{0}$. Consequently, terms $g(A,B,C,D)$ and $h(A,B,C,D)$ do not matter here, and we only need the terms 
$(A-BD^{-1}C)^{-1}$ and $-(A-BD^{-1}C)^{-1}BD^{-1}$. 
The first term $(A-BD^{-1}C)^{-1}$ corresponds to 
\begin{flalign*}
\scriptsize
\setlength{\arraycolsep}{1pt}
\Big(
\begin{bmatrix}
\mathcal{M} & \mathcal{J}^T_e (x_t) \\
\mathcal{J}_e (x_t)& \mathbf{0} \\ 
\end{bmatrix}
\hspace{-0.1cm} -\hspace{-0.1cm} 
\begin{bmatrix}
\mathcal{J}^T_f(\dot{x}_t)  & \mathbf{0}\\
\mathbf{0} & \mathbf{0} \\ 
\end{bmatrix}
\hspace{-0.1cm} 
\begin{bmatrix}
D(-\mathcal{J}_f(\dot{x}_t)  \dot{x}^*_{t+1} - \gamma^*) & -D(\lambda_f^*)\\
D(\gamma^*) & \hspace{-1cm}D(\lambda^*_f-\mu\diag(\mathcal M)) \\ 
\end{bmatrix}^{-1}
\hspace{-0.1cm} 
\\ \hspace{-0.1cm} 
\scriptsize
\begin{bmatrix}
D(\lambda^*)\mathcal{J}_f(\dot{x}_t) & \mathbf{0}\\
\mathbf{0} & \mathbf{0} \\ 
\end{bmatrix}
\Big)^{-1}.
\end{flalign*}
In the model identification phase, we only utilize data points where the object actually moves when pushed by the robot. Thus, $-\mathcal{J}_f(\dot{x}_t)  \dot{x}^*_{t+1} - \gamma^*= \mathbf{0}$ and $\lambda^*_f-\mu\diag(\mathcal M) = \mathbf{0}$, and 
\begin{eqnarray*}
\footnotesize
\setlength{\arraycolsep}{5pt}
(A-BD^{-1}C)^{-1} = 
\begin{bmatrix}
\mathcal{M} & \mathcal{J}^T_e (x_t) \\
\mathcal{J}_e (x_t)& \mathbf{0} \\ 
\end{bmatrix}^{-1}
= 
\begin{bmatrix}
X_{1,1} & X_{1,2} \\
X_{2,1} & X_{2,2} \\
\end{bmatrix}.
\end{eqnarray*}
Using the blockwise matrix inversion, we find that 
{\small
$$X_{1,1} = \mathcal{M}^{-1} +  \mathcal{M}^{-1} \mathcal{J}^{T}_e(x_t) (-\mathcal{J}_e(x_t) \mathcal{M}^{-1} \mathcal{J}^{T}_e(x_t))^{-1} \mathcal{J}_e(x_t)  \mathcal{M}^{-1}.$$ }We will see in the following that the remaining matrices, $X_{1,2}, X_{2,1}$ and $X_{2,2}$, will not be needed. 
Similarly, the top right of $G^{-1}$ is $-(A-BD^{-1}C)^{-1}BD^{-1}$. It is given as 
\begin{flalign*}
\scriptsize
\setlength{\arraycolsep}{1pt}
-\Big(
\begin{bmatrix}
X_{1,1} & X_{1,2} \\
X_{2,1} & X_{2,2} \\
\end{bmatrix}
\begin{bmatrix}
\mathcal{J}^T_f(\dot{x}_t) & \mathbf{0}\\
\mathbf{0} & \mathbf{0} \\
\end{bmatrix}
\begin{bmatrix}
\mathbf{0} & D^{-1}(\gamma^*)\\
-D^{-1}(\lambda^*_f) & \mathbf{0} \\
\end{bmatrix}
\Big)
\\
\scriptsize
=
\begin{bmatrix}
\mathbf{0} & X_{1,1} \mathcal{J}^T_f(\dot{x}_t)  D^{-1}(\gamma^*)\\
\mathbf{0} & X_{2,1} \mathcal{J}^T_f(\dot{x}_t)  D^{-1}(\gamma^*)\\
\end{bmatrix}
.
\end{flalign*}
Therefore,
\begin{eqnarray*}
\footnotesize
\setlength{\arraycolsep}{5pt}
G^{-1} = 
\begin{bmatrix}
X_{1,1} & X_{1,2} & \mathbf{0} & X_{1,1} \mathcal{J}^T_f(\dot{x}_t)  D^{-1}(\gamma^*)\\
X_{2,1} & X_{2,2} & \mathbf{0} & X_{2,1} \mathcal{J}^T_f(\dot{x}_t)  D^{-1}(\gamma^*)\\
X_{3,1} & X_{3,2} & X_{3,3} & X_{3,4}\\
X_{4,1} & X_{4,2} & X_{4,3} & X_{4,4}
\end{bmatrix}.
\end{eqnarray*}
The first term in $[\alpha_1,\alpha_2,\alpha_3,\alpha_4]$ is then {\footnotesize $$\alpha_1 = \frac{\partial loss}{\partial (-\dot{x}_{t+dt})} X_{1,1},$$}and the fourth term is 
{\footnotesize$$\alpha_4 =  \frac{\partial loss}{\partial (-\dot{x}_{t+1})} \big(-X_{1,1}  \mathcal{J}^T_f(\dot{x}_t)  D^{-1}(\gamma^*)\big).$$}Since
$
\frac{\partial loss}{\partial (-\dot{x}_{t+1})} (-\partial \dot{x}_{t+1}) = [\alpha_1,\alpha_2,\alpha_3,\alpha_4] [\partial  \mathcal M (\dot{x}^*_{t+1} - \dot{x}_{t}), \mathbf{0} , \mathbf{0}, D(\gamma^*)\partial (\mu\diag(\mathcal M))]^T
$,
then 
{\footnotesize
\begin{eqnarray}
\frac{\partial loss}{\partial (-\dot{x}_{t+1})} (-\partial\dot{x}_{t+1}) =  \frac{\partial loss}{\partial (-\dot{x}_{t+1})}  \Big( X_{1,1}  \partial  \mathcal M (\dot{x}^*_{t+1} - \dot{x}_{t})  \nonumber \\
+ \big(-X_{1,1}  \mathcal{J}^T_f(\dot{x}_t)  D^{-1}(\gamma^*)\big) D(\gamma^*)\partial \big(\mu\diag(\mathcal M)\big)  \Big)  \nonumber \\
=  \frac{\partial loss}{\partial (-\dot{x}_{t+1})}  \Big( X_{1,1}  \partial  \mathcal M (\dot{x}^*_{t+1} - \dot{x}_{t}) + \big(-X_{1,1}  \mathcal{J}^T_f(\dot{x}_t)\big) \partial \big(\mu\diag(\mathcal M)\big)  \Big)  \nonumber 
\label{velocityGradient}
\end{eqnarray}}
In the following, we show how to use the equation above to derive $\frac{\partial loss}{\partial \mathcal{M}}$ and $\frac{\partial loss}{\partial \mu}$ and use them in a {\it coordinate descent} algorithm to identify $(\mathcal{M}^*,\mu^*)$ from data.
\subsection{Mass Gradient}
We calculate~$\frac{\partial loss}{\partial \mathcal{M}}$ while setting $\partial \mu = \mathbf{0}$. 
{\footnotesize
\begin{eqnarray*}
\frac{\partial loss}{\partial (-\dot{x}_{t+1})} (-\partial\dot{x}_{t+1}) =  \frac{\partial loss}{\partial (-\dot{x}_{t+1})}  \Big( X_{1,1}  \partial  \mathcal M (\dot{x}^*_{t+1} - \dot{x}_{t}) 
\\ + \big(-X_{1,1}  \mathcal{J}^T_f(\dot{x}_t)\big) \mu \partial \diag(\mathcal M)  \Big)  \nonumber 
\label{velocityGradient}
\end{eqnarray*}}
Then,
{\footnotesize
\begin{eqnarray}
\frac{\partial loss}{\partial (-\dot{x}_{t+1})} \frac{(-\partial \dot{x}_{t+1})}{\partial \mathcal{M}} 
=  \frac{\partial loss}{\partial (\dot{x}_{t+1})}  X_{1,1}  \big(\mathcal{J}^T_f(\dot{x}_t)\mu - D(\dot{x}^*_{t+1}- \dot{x}_{t})\big) \nonumber 
\end{eqnarray}}
From the definition of the loss function in Equation~\ref{simulationError}, we can see that $\frac{\partial loss}{\partial (\dot{x}_{t+1})} = 2dt \sum_{t=1}^{T-1} D\big(x_{t+1} - x^g_{t+1}\big)$, wherein $x_{t+1}^g$ is the observed ground-truth pose 
of the object and $x_{t+1}$ is its predicted pose, computed as $x_{t+1} = x^g_{t}+\dot{x}^*_{t}dt$. Finally,
{\footnotesize
\begin{eqnarray*}
\frac{\partial loss}{\partial \mathcal{M}} = 2dt \sum_{t=1}^{T-1}\Big( D\big(x_{t+1} - x^g_{t+1}\big)  X_{1,1} \big(\mathcal{J}^T_f(\dot{x}_t)\mu - D(\dot{x}^*_{t+1}- \dot{x}_{t})\big)\Big).
\label{massGradient}
\end{eqnarray*}}

\subsection{Friction Gradient}
We calculate~$\frac{\partial loss}{\partial \mu}$ while setting $\partial \mathcal{M} = \mathbf{0}$. 
{\footnotesize
\begin{eqnarray*}
\frac{\partial loss}{\partial (-\dot{x}_{t+1})} (-\partial\dot{x}_{t+1}) =  \frac{\partial loss}{\partial (-\dot{x}_{t+1})}  \big(-X_{1,1}  \mathcal{J}^T_f(\dot{x}_t) \partial \mu \diag(\mathcal M) \big) 
\end{eqnarray*}}
Then,
{\footnotesize
\begin{eqnarray*}
\frac{\partial loss}{\partial (-\dot{x}_{t+1})} \frac{(-\partial \dot{x}_{t+1})}{\partial \mathcal{\mu}} =  \frac{\partial loss}{\partial (-\dot{x}_{t+1})}  \big(-X_{1,1}  \mathcal{J}^T_f(\dot{x}_t) \mathcal M\big) 
\end{eqnarray*}}
Finally,
{\footnotesize
\begin{eqnarray*}
\frac{\partial loss}{\partial \mu} =  2dt  \sum_{t=1}^{T-1} \Big( D\big(x_{t+1} - x^g_{t+1}\big) X_{1,1}  \mathcal{J}^T_f(\dot{x}_t) \mathcal M \Big) 
\end{eqnarray*}}

\subsection{Mass and Friction Identification Algorithm}
  \begin{algorithm}[h]
{\small
     \SetAlgoLined
\KwIn{
A set of real-world trajectories $\mathcal D = \{ (x^g_{t},\dot{x}^g_{t},F_{t})_{t=0}^{T} \}$;
a learning rate $\alpha_{\textrm{rate}}$; and initial mass matrix $\mathcal{M}$ and friction map $\mu$;
Maximum mass and friction $m_{max},u_{max}\in \mathbb{R}$ \;
}
\KwOut{Updated mass matrix $\mathcal{M}$ and friction map $\mu$}

\Repeat{timeout}{
\ForEach{$\mathcal T^{g}  =  (x^g_{t},\dot{x}^g_{t},F_{t})_{t=0}^{T}  \in \mathcal D$}
{
$(x_0,\dot{x}_0) \leftarrow (x^g_0,\dot{x}^g_0)$;$x_1 \leftarrow x^g_1$\;
\For{$t\in\{0,T-2\}$}{
   $\dot{x}_{t+1} \leftarrow V(x_{t},\dot{x}_{t}, F_t, \mathcal M, \mu) $ \Comment*[r]{\tiny \textcolor{blue}{Solving the LCP}}
   $x_{t+2} \leftarrow x_{t+1} + \dot{x}_{t+1}dt$\Comment*[r]{\tiny \textcolor{blue}{Predicting next pose}}
  {\footnotesize
  $X\leftarrow \mathcal{M}^{-1} +  \mathcal{M}^{-1} \mathcal{J}^{T}_e(x_t) (-\mathcal{J}_e(x_t) \mathcal{M}^{-1} \mathcal{J}^{T}_e(x_t))^{-1} \mathcal{J}_e(x_t)  \mathcal{M}^{-1}$}
   $W_{\mu} \leftarrow X \mathcal{J}^T_f(\dot{x}_t) \mathcal M$\;
  $W_{\mathcal{M}} \leftarrow X \big(\mathcal{J}^T_f(\dot{x}_t)\mu - D(\dot{x}_{t+1}- \dot{x}_{t})\big)$\;
   $\mu \leftarrow \mu - \alpha_{\textrm{rate}} D\big(x_{t+1} - x^g_{t+1}\big) W_{\mu}$\;
  $\mathcal{M} \leftarrow \mathcal{M} - \alpha_{\textrm{rate}} D\big(x_{t+1} - x^g_{t+1}\big) W_{\mathcal M}$\;
  $\mu \leftarrow \arg\min_{\mu'\in[\mathbf{0}, I u_{max}]} \|\mu'-\mu\|_\infty$\;
  $\mathcal{M} \leftarrow \arg\min_{\mathcal{M}'\in]\mathbf{0}, I m_{max}]} \|\mathcal{M}'-\mathcal{M}\|_\infty$
    \Comment*[r]{\tiny \textcolor{blue}{Projecting the gradients}}
}
}
}
\caption{Learning Mass and Friction with Differentiable Physics Simulations}
\label{identificationAlgo}
}
\end{algorithm}
The gradient of the loss function with respect to the mass and friction matrices $\mathcal M$ and $\mu$ are used in Algorithm~\ref{identificationAlgo} to search for the ground-truth mass and friction.
We present here the stochastic version where the gradient is computed for each trajectory separately. The gradient can also be computed in a batch mode from all trajectories. 
$\{(x^g_{t},\dot{x}^g_{t},F_{t})_{t=0}^{T}\}$ of data collected by the robot. The mass and friction are increased or decreased depending on the signs in the error vector $\big(x_{t+1} - x^g_{t+1}\big)$, which corresponds to 
the reality gap. The main computational bottleneck is in computing  weights $W_{\mathcal{M}}$ and $W_{\mu}$, which is linear in the number of cells $n$ because $\mathcal M$ is a diagonal matrix, and $\mathcal{J}_e(x_t)$ is block diagonal thanks to 
 the regular grid structure of the cells.
Finally, we project the updated mass and friction matrices by rounding their values down to upper limits $m_{max},u_{max}\in \mathbb{R}$. 
The provided upper limits are the same for every cell in the object, whereas the mass and friction distributions learned by the algorithm are highly heterogenous, as will be shown in the experiments. 

\section{Policy Gradient}
After identifying the mass distribution $\mathcal M$ and friction map $\mu$ using Algorithm~\ref{identificationAlgo}, we search for a new sequence of forces $(F_t)_{t=0}^{T-1}$ to push the object toward a desired terminal goal configuration 
$x_T^d$. Algorithm~\ref{policyAlgo} summarizes the main steps of this process. We start by creating a rapidly exploring random tree ($RRT^*$) to find the shortest path from $x_0$ to $x_T^d$. 
While searching for the shortest path, we eliminate from the tree object poses that are unstable (based on the identified mass distribution $\mathcal M$) or that are in collision with other objects. 
$RRT^*$ returns a set of waypoints $\mathcal X_{waypoint}$. At each iteration of the main loop of the algorithm, we find the nearest waypoint in $\mathcal X_{waypoint}$ and search for actions that would push the object toward it.
A  pushing force is parameterized by a contact point, a direction and a magnitude, as discussed in Section~\ref{problem_setup}. We focus here on optimizing the contact point, and we keep the magnitude constant. 
The direction of the force is chosen to be always horizontal. It is given as the opposite of the surface normal of the object at the contact point, projected down on the 2D plane of the support surface. 
This choice is made to avoid slippages and changes in contact points during a push. 

A contact point is always located on the outer side of a cuboid (cell). Therefore, we limit the search to the outer cells of the grid. The objective of this search is to select a contact point that reduces the gap between the predicted
pose of the object after pushing it, and the nearest waypoint $x_{target}$ that has not been reached yet. 
We select the initial contact point as the outer cell that is most aligned to the axis $\hat{x} - x_{target}$, where $\hat{x}$ is the estimated center of mass.
The gradient of the gap with respect to the contact point is computed by using the finite-difference method. The contact point is moved in the direction that minimizes the gap until a local optimum is reached. 
Force $F_t$ is then defined based on the selected contact point. The pose and velocity of the object are replaced by the predicted ones that result from applying force $F_t$. This process is repeated until the object reaches the desired 
goal configuration. The time duration of each pushing action is also optimized by using finite differences. This part is omitted for simplicity's sake.

The surface of the object often contains non-differentiable parts where the analytical gradient with respect to the contact point is undefined.  Even on smooth parts, there is no clear advantage of computing the analytical gradient here, because the space of contact points is uni-dimensional, in contrast to the high-dimensional space of non-uniform mass and friction distributions. In low-dimensional search spaces, finite-difference methods are computationally efficient. 


  \begin{algorithm}[h]
{  \small
     \SetAlgoLined
\KwIn{Identified mass and friction $\mathcal M$ and $\mu$, and presumed $3D$ shape $\mathcal S$ of the object; initial ground-truth configuration $x_0$; desired final configuration $x_T^d$;}
\KwOut{A sequence of pushing forces $(F_t)_{t=0}^{T-1}$ ; }
Find a set $\mathcal X_{waypoint}$ of waypoints by calling $RRT^*(\mathcal S, x_0, x_T^d)$\;
$\dot{x}_0 \leftarrow \mathbf{0}; t \leftarrow 0 $\;
\While{$ \mathcal X_{waypoint} \neq \emptyset $}
{
\Repeat{$\min_{x\in \mathcal X_{waypoint}} \|x- x_t\|_2 \leq \epsilon$}{
   $\hat{x} \leftarrow \frac{\sum_{i\in \{1,\dots,n\}}  (\mathcal M_i) (p^i_{x,t},p^i_{y,t})}{\sum_{i\in \{1,\dots,n\}} \mathcal M_i}$    \Comment*[r]{\tiny \textcolor{blue}{center of mass}}
   {$x_{target} \leftarrow \arg\min_{x\in \mathcal X_{waypoint}} \|x- x_t\|_2$}\;
   {\small $i^* \leftarrow \arg\max_{i\in \{1,\dots,n\}} \frac{ \big(\hat{x} - x_{target}\big)\big(  (p^i_{x,t},p^i_{y,t}) - \hat{x} \big)^T }{\| \hat{x} - x_{target}  \|_2   \|(p^i_{x,t},p^i_{y,t}) - \hat{x} \|_2 } $} \Comment*[r]{\tiny \textcolor{blue}{Initializing the contact point}}
   improvement $\leftarrow$ true\;
   \Repeat{improvement = false}{
   {\footnotesize Let $i^*_{\textrm{left}}$ and $i^*_{\textrm{right}}$ be two cells that are adjacent to $i^*$ and that are also on the outer envelope of the object\;}
   \For{$i\in\{i^*,i^*_{\textrm{left}},i^*_{\textrm{right}}\}$}
   {
      {\footnotesize Choose the direction of the force $(f^i_{x},f^i_{y})$ as the opposite of the object' surface normal at cell $i$.}
   $F^i \leftarrow [0,0,0,\dots,f^i_{x},f^i_{y},0,\dots, 0,0,0]$\;
      $gap_{i} \leftarrow \| x_{t} + \dot{x}_{t}dt + V(x_{t},\dot{x}_{t}, F^i, \mathcal M, \mu)dt - x_{target} \|_2 $\;
   }
   $i^*\leftarrow \arg\min_{i \in\{i^*,i^*_{\textrm{left}},i^*_{\textrm{right}}\}} gap_i$\;
   $F_t \leftarrow [0,0,0,\dots,f^{i^*}_{x},f^{i^*}_{y},0,\dots, 0,0,0] $\;
   $\dot{x}_{t+1} \leftarrow V(x_{t},\dot{x}_{t}, F^{i^*}, \mathcal M, \mu)$; ${x}_{t+1} \leftarrow {x}_t + {\dot{x}_t}dt$\;
   \If{$i^* \neq i^*_{\textrm{left}} \wedge i^* \neq i^*_{\textrm{right}}$}{   improvement $\leftarrow$ false\;}
   }
$t\leftarrow t+1$\;
}
{\small
$\mathcal X_{waypoint} \leftarrow \mathcal X_{waypoint} \backslash  \{\arg\min_{x\in \mathcal X_{waypoint}} \|x- x_t\|_2\}$}\;
}
\caption{Planning and Control}
\label{policyAlgo}
}
\end{algorithm}

\section{Main Algorithm}
Algorithm~\ref{mainAlgo} summarizes the main steps of the proposed approach and the protocol followed in the experiments. In summary, the robot first ``plays'' with the unknown object by
applying random short-lasting horizontal forces for a safe and local exploration. A mass and friction model is then inferred from the gathered data by using Algorithm~\ref{identificationAlgo}. Based on 
the inferred mass distribution, a safe goal configuration is sampled from a desired goal region. For example,
a pregrasp sliding manipulation can be used to grasp a thin object that cannot be directly grasped from a flat surface. 
In~\cite{Kaiyu2019}, a known object is pushed to the edge of a table and then grasped from there. 
Pushing an unknown object to the edge of a table results often in losing the object. 
Our method avoids this issue by sampling a goal configuration that allows a sufficient part of the object to be graspable, while keeping the object balanced on the edge thanks to the identified mass distribution. 
Once a goal is selected, Algorithm~\ref{policyAlgo} is used to generate a sequence of actions to push the object to the goal.

  \begin{algorithm}[h]
  {  \small
     \SetAlgoLined
\KwIn{Point cloud of an object's upper surface; Maximum mass and friction $m_{max},u_{max}\in \mathbb{R}$; Desired goal region $\mathcal G$;}
Create a 3D shape $\mathcal S$ of the object by projecting its upper surface down on the support surface\; 
Decompose the 3D shape into a regular grid of $n$ small cuboids\;  
Let $x^g_0$ be the vector of ground-truth positions and rotations of the $n$ cuboids at time $0$; $\dot{x}^g_0 = \mathbf{0}$\;
Initialize mass matrix $\mathcal M$ and friction map $\mu$ uniformly with a  random value\;
Sample a small number $T_{\textrm{\tiny exploration}}$ of contact points and horizontal pushing directions \Comment*[r]{\tiny \textcolor{blue}{For a safe local exploration}}
Use the robot's end-effector to apply forces on the sampled contact points and along the sampled directions for short periods of time $dt$, 
and record the resulting trajectories in $\mathcal D = \{ (x^g_{t},\dot{x}^g_{t},F_{t})_{t=0}^{T_{\textrm{\tiny exploration}}} \}$\;
Use Algorithm~\ref{identificationAlgo} to identify  $\mathcal M$  and $\mu$ from dataset $\mathcal D$ \;
Sample a goal configuration ${x}^d \in \mathcal G$ where the object is predicted to remain stable under gravity, according to the identified mass distribution $\mathcal M$\;
Use Algorithm~\ref{policyAlgo} to obtain a new sequence of forces $(F_{t})_{t=0}^{T_{\textrm{\tiny execution}}}$ to push the object to ${x}^d$\;
Execute the sequence $(F_{t})_{t=0}^{T_{\textrm{\tiny execution}}}$ with the robot's end-effector\;
\caption{Learning to Slide Unknown Objects with Differentiable Physics Simulations}
\label{mainAlgo}
}
\end{algorithm}

\section{Evaluation}
We report here the results of three sets of experiments to evaluate the proposed approach. The most important set is the one related to mass and friction identification.
\subsection{Experimental Setup}
The experiments are performed on both simulated and real robot and objects.
In the real robot setup, a {\it Robotiq} 3-finger hand mounted on a {\it Kuka} robot is repeatedly moved to collide with a rigid object that is set on a table-top and to push it forward, as shown as Figure \ref{fig:robot}. 
The initial, final and intermediate point clouds of the object are recorded.
The simulation experiments are performed using the physics engine \textit{Bullet} and models of the robot and objects.
The experiments are performed on five real objects: a book, a hammer, a snack, a toolbox and a spray gun, and eight simulated objects: a box, a hammer, a book, a crimp, a snack, a ranch, a spray gun and a toothpaste.
The number of cells per object varies from $28$ to $88$ depending on the size of the object.

\subsection{Tasks}
{\bf Model Identification}. Each real and simulated object is pushed by the robot randomly $10$ times on the table. 
Half of the recorded trajectories are used for learning a mass matrix $\mathcal M$ and friction map $\mu$. The other half is used for testing the identified models. 
Since the ground-truth values of mass and friction are unknown, the identified models are evaluated in terms of the accuracy of the predicted pose of each cell after applying the sequence of actions provided in the test set.
The experiments on the real objects are repeated with $25$ randomized splits into training and testing sets. The simulation experiments are repeated with $10$ different models per object. 

{{\bf Planning and Control}. For each one of the eight objects in simulation, we randomly sample $10$ values for their mass and friction matrices, 
and $10$ random goal configurations in a disk of a radius of $1m$ around the initial configuration. The rotations of the goal configurations are also selected randomly. 
The number of settings is then $8\times10\times10$. The task is to generate, in each setting, a sequence of forces that pushes the object from the initial configuration to the goal.
The objective is to asses the computational efficiency of Algorithm~\ref{policyAlgo}. 

{{\bf Pre-grasp Sliding Manipulation}. Finally, we evaluate the entire system (Algorithm~\ref{mainAlgo}) on the task of sliding an object 
 from a random initial pose on a table to a desired goal region at the edge of the table where the object can be grasped.
The object cannot be directly grasped from a flat surface, the goal is to push it to the edge where part of it sticks out of the table and becomes graspable.
Since the mass distribution of the object is highly heterogeneous and unknown, 
the object often becomes unbalanced at the edge and falls from the table if the identified model is incorrect.
We report here the percentage of experiments where the object is successfully pushed to the edge and grasped without losing it.
The experiments on this task are performed using the real Kuka robot and a real hammer.  
The exploration phase contains only $5$ random pushing actions that are used for model identification. 
The reported results are averaged over $16$ independent runs, with a different initial pose in each run. 



\begin{figure}[ht]
    \centering
    \begin{tabular}{cc}
        \includegraphics[width=0.15\textwidth]{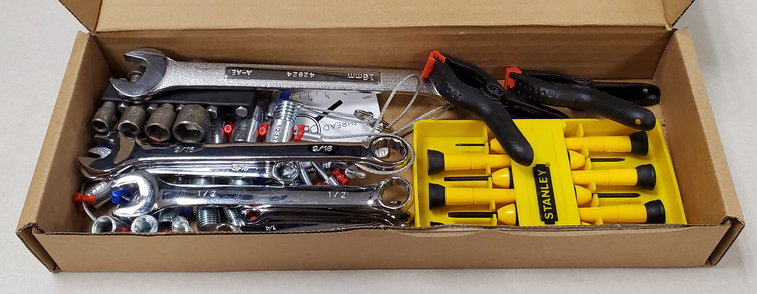} & 
        \includegraphics[width=0.15\textwidth]{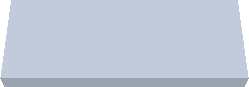}
        \includegraphics[width=0.15\textwidth]{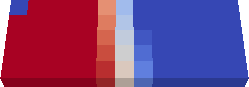}
        \\
        \includegraphics[width=0.15\textwidth]{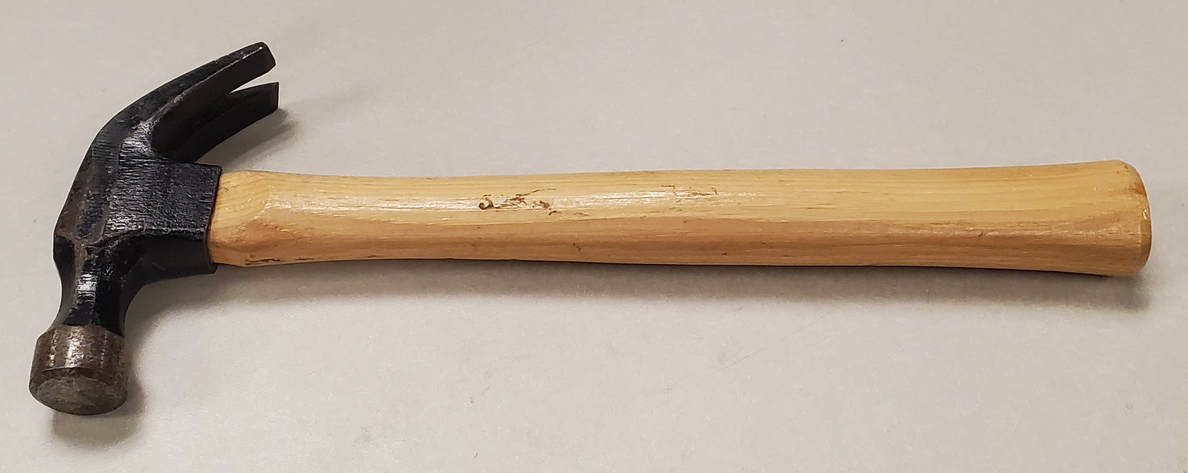} &  
        \includegraphics[width=0.15\textwidth]{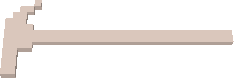}
        \includegraphics[width=0.15\textwidth]{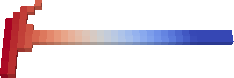}
        \\
        \includegraphics[width=0.15\textwidth]{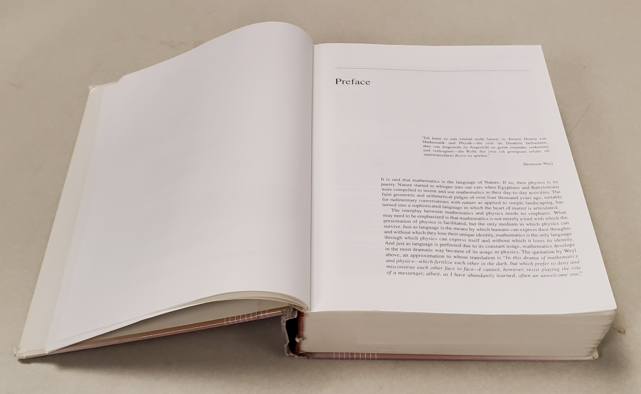} & 
        \includegraphics[width=0.15\textwidth]{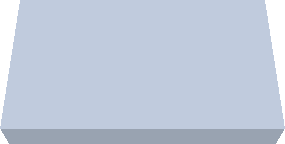}
        \includegraphics[width=0.15\textwidth]{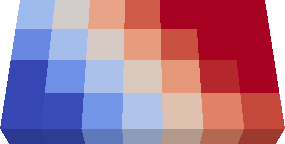}
        \\
        \includegraphics[width=0.15\textwidth]{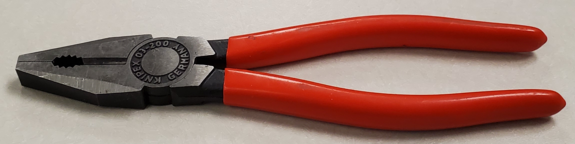} & 
        \includegraphics[width=0.15\textwidth]{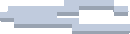}
        \includegraphics[width=0.15\textwidth]{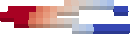}
        \\
        \includegraphics[width=0.15\textwidth]{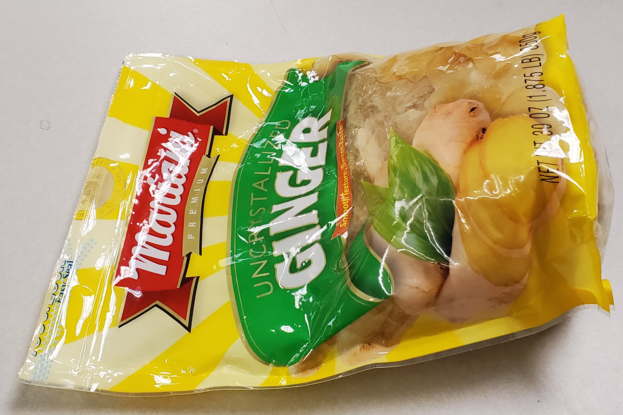} & 
        \includegraphics[width=0.15\textwidth]{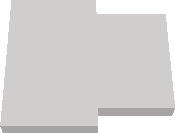}
        \includegraphics[width=0.15\textwidth]{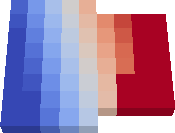}
        \\
        \includegraphics[width=0.15\textwidth]{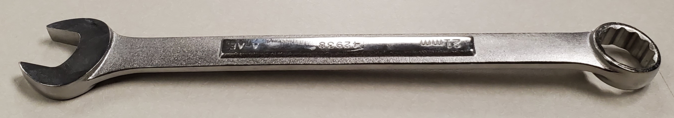} & 
       \includegraphics[width=0.15\textwidth]{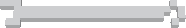}
        \includegraphics[width=0.15\textwidth]{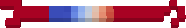}
        \\
        \includegraphics[width=0.15\textwidth]{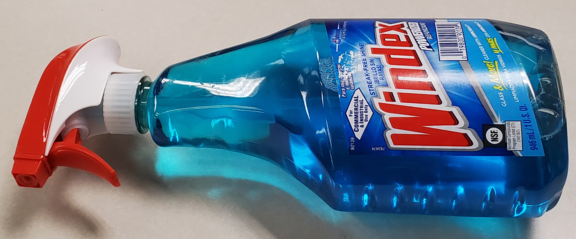} & 
        \includegraphics[width=0.15\textwidth]{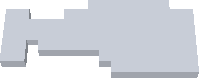}
        \includegraphics[width=0.15\textwidth]{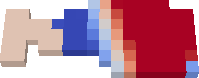}
        \\        
        \includegraphics[width=0.15\textwidth]{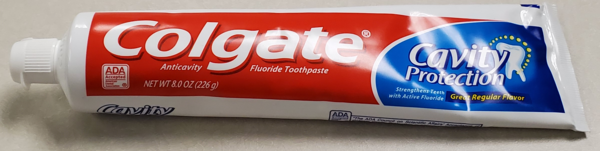} & 
        \includegraphics[width=0.15\textwidth]{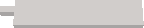}
        \includegraphics[width=0.15\textwidth]{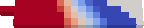}
    \end{tabular}
    \caption{\small Learned mass distributions. Red color means higher mass value while blue color means lower mass value. 
    The middle column shows the initial uniform mass distributions provided to Algorithm~\ref{identificationAlgo} and the right column shows the mass distributions returned by the algorithm after only five iterations of gradient descent.}
    \label{fig:qual_heatmap}
\end{figure}

\vspace{-0.3cm}
\subsection{Compared Methods}

{\bf Model Identification}. Algorithm~\ref{identificationAlgo} is compared against the following methods. \emph{Random search} is a baseline method that repeatedly samples random values of the mass and friction matrices and returns the best sampled values that minimize $loss(\mathcal M, \mu)$.
\emph{Weighted sampling search} generates random values uniformly in the first iteration, and then iteratively generates normally distributed random values around the best parameter obtained in the previous iteration.
The standard deviation of the random values is gradually reduced over time, to focus the search on the most promising region of the search space.
\emph{The finite differences gradient} is an approximation of the analytical gradient.
We add or subtract a small amount to the current parameter values and simulate the trajectories using the neighboring parameter values to approximate the derivatives $\frac{\partial loss}{\partial \mathcal M}$ and $\frac{\partial loss}{\partial \mu}$.
Because a large number of simulations is required to compute the gradient for all the cells, the parameters of each cell are updated using the coordinate gradient descent.
We also compare the proposed method with two \textit{black-box} optimization methods: \textit{CMA-ES}~\cite{hansen2006cma} and \textit{Nelder-Mead}, and an automatic differentiation of the LCP solver using the \textit{Autograd} function of PyTorch \cite{paszke2017automatic}.
The same minimum and maximum bounds of mass and friction are provided to all methods and are also used for all cells of objects.

{{\bf Planning and Control}. We perform an ablation study where we substitute the finite-difference gradient in Algorithm~\ref{policyAlgo} with an \emph{exhaustive search} of the optimal contact point.

{{\bf Pre-grasp Sliding Manipulation}. We compare  Algorithm~\ref{mainAlgo} to two alternatives. The first one assumes a uniform and homogenous mass and friction values, and uses directly  Algorithm~\ref{policyAlgo} to push the object to the goal \emph{without model identification}. The second alternative is identical to Algorithm~\ref{mainAlgo}, except that \emph{no upper limit} on the friction coefficients is used.

\begin{figure*}[h]
	\begin{tabular}{@{\hskip3pt}c@{\hskip3pt}c@{\hskip3pt}c@{\hskip3pt}c@{\hskip3pt}}
 		\includegraphics[width=0.242\textwidth]{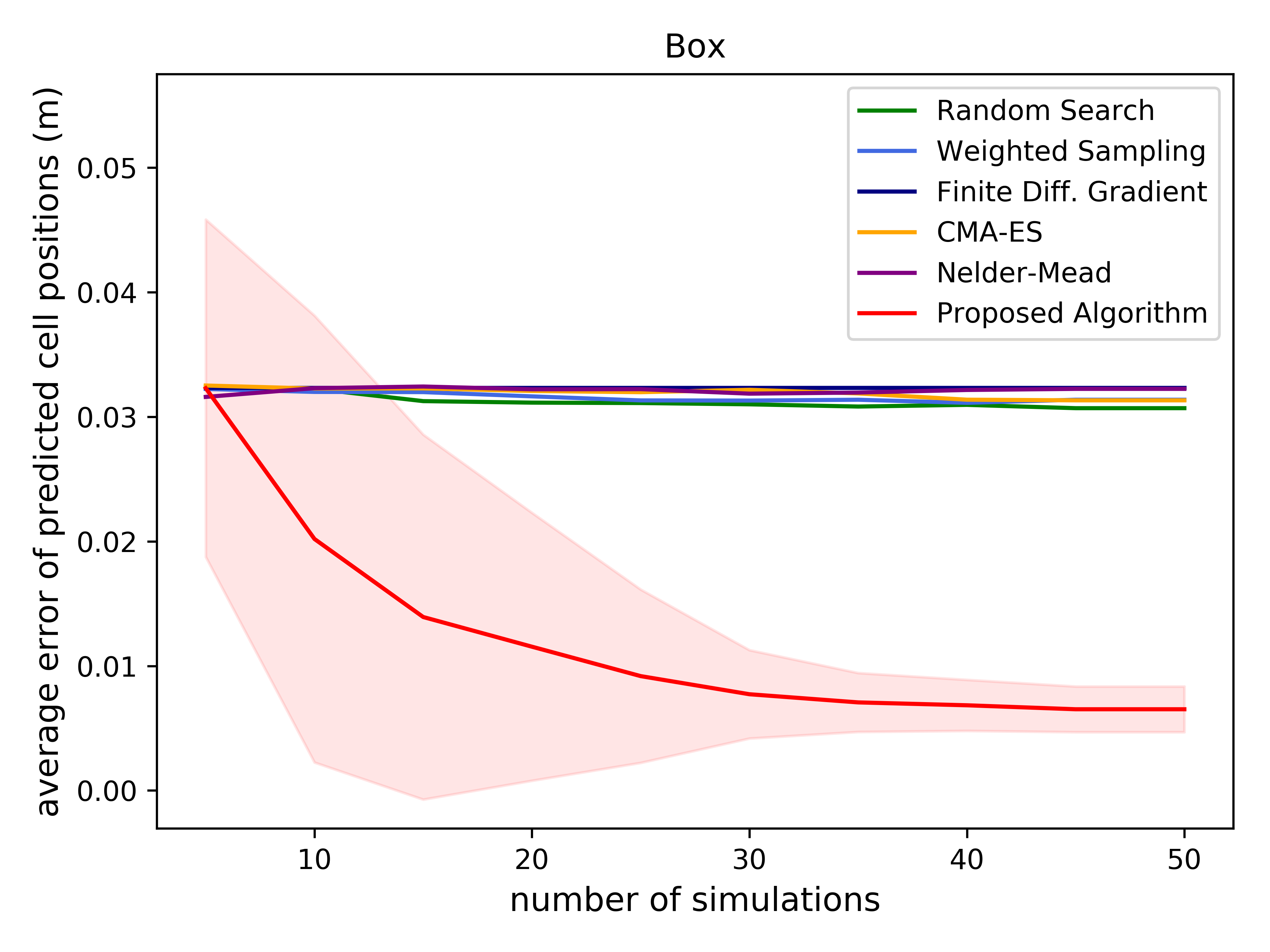} &
 		\includegraphics[width=0.242\textwidth]{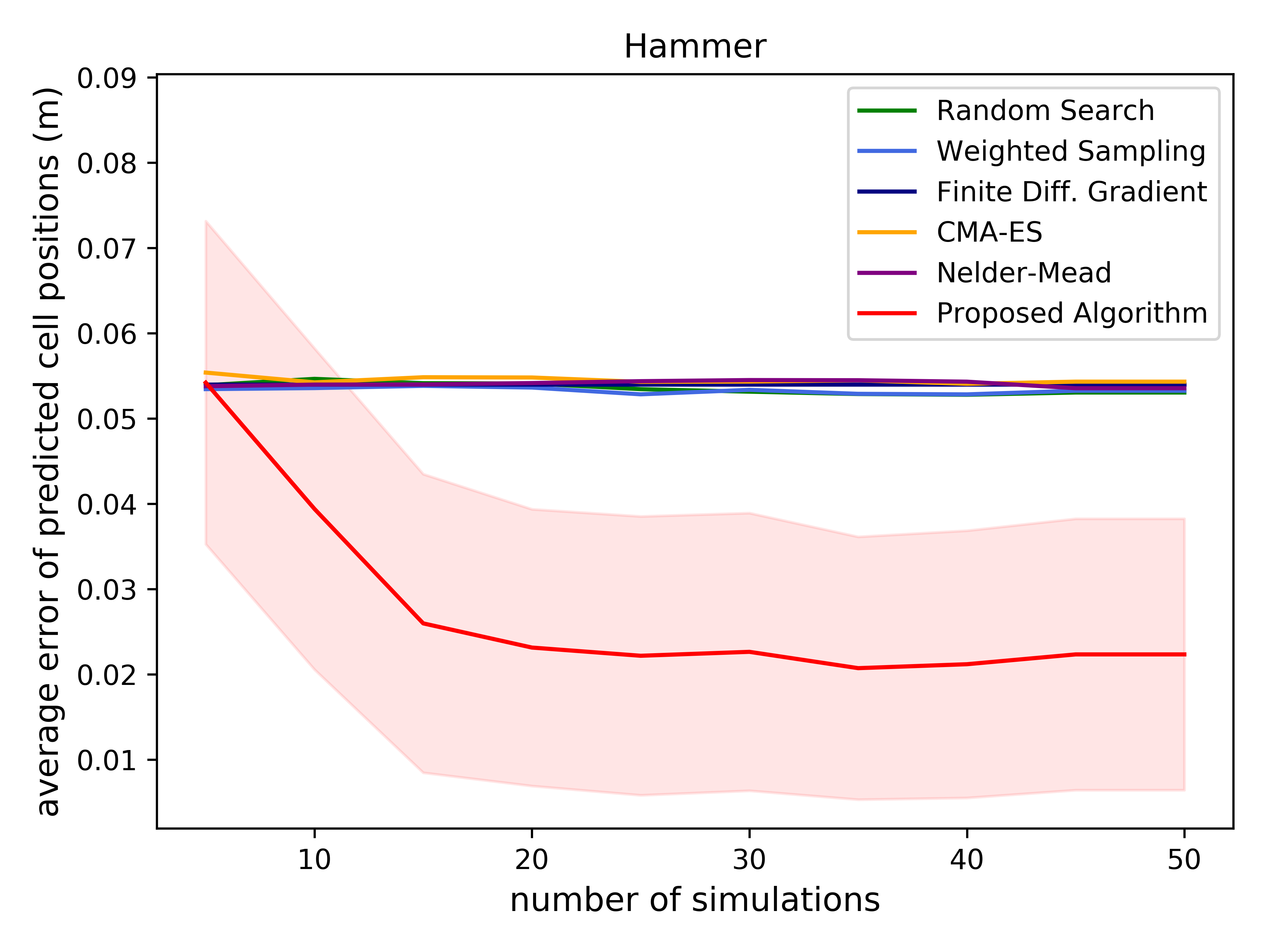} &
 		\includegraphics[width=0.242\textwidth]{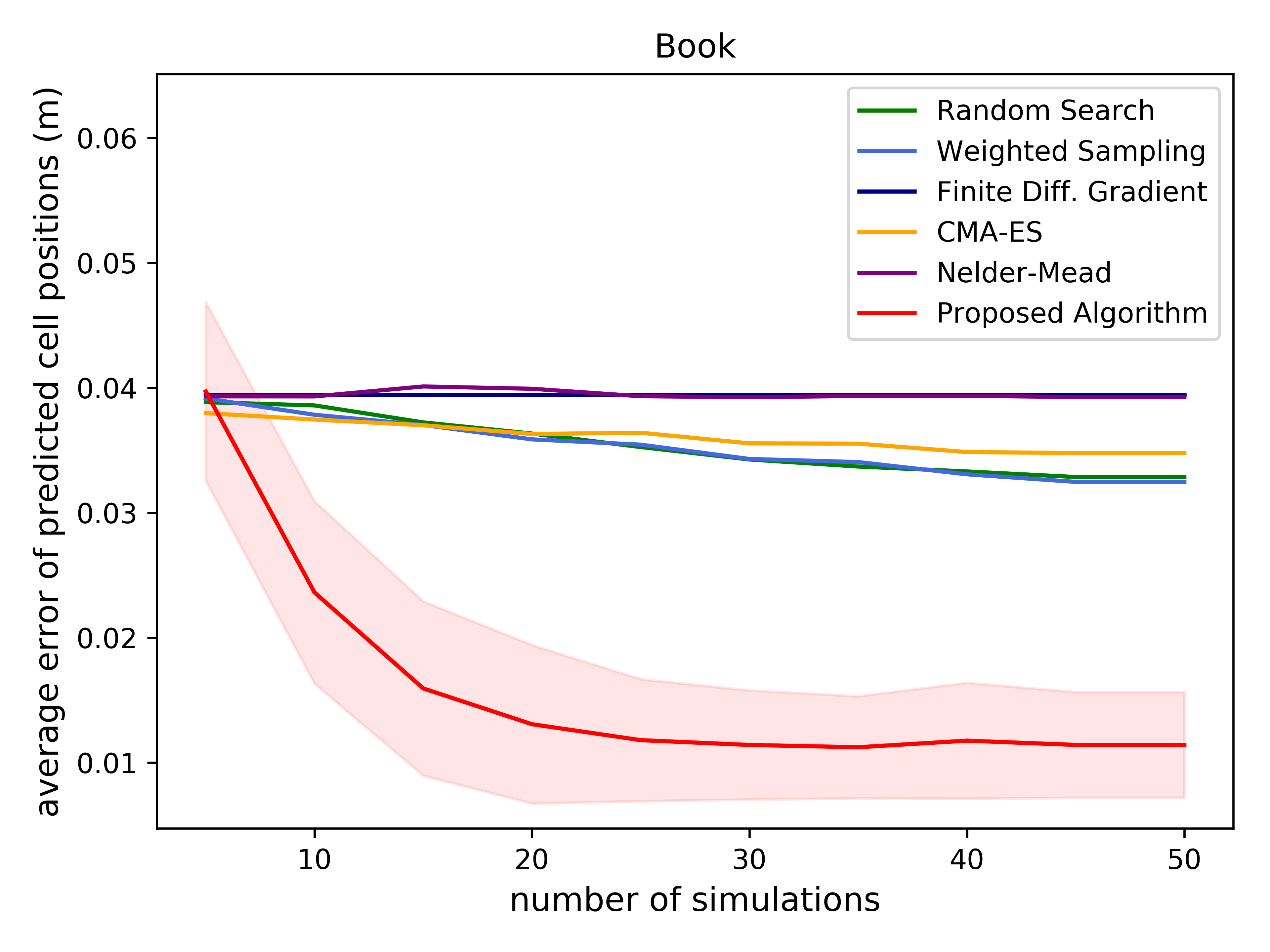} &
 		\includegraphics[width=0.242\textwidth]{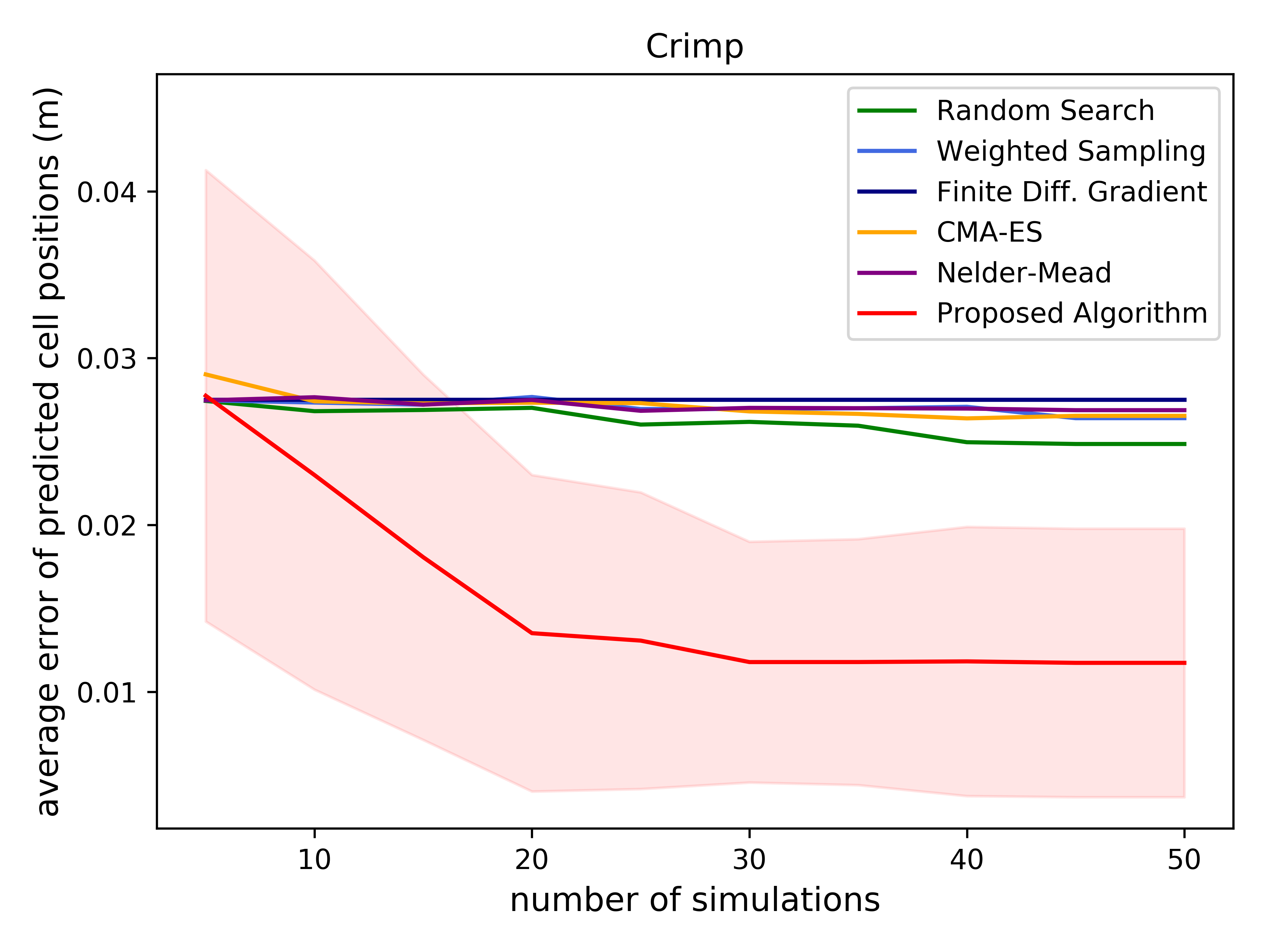} \\
 		(a) & (b) & (c) & (d)\\
		\includegraphics[width=0.242\textwidth]{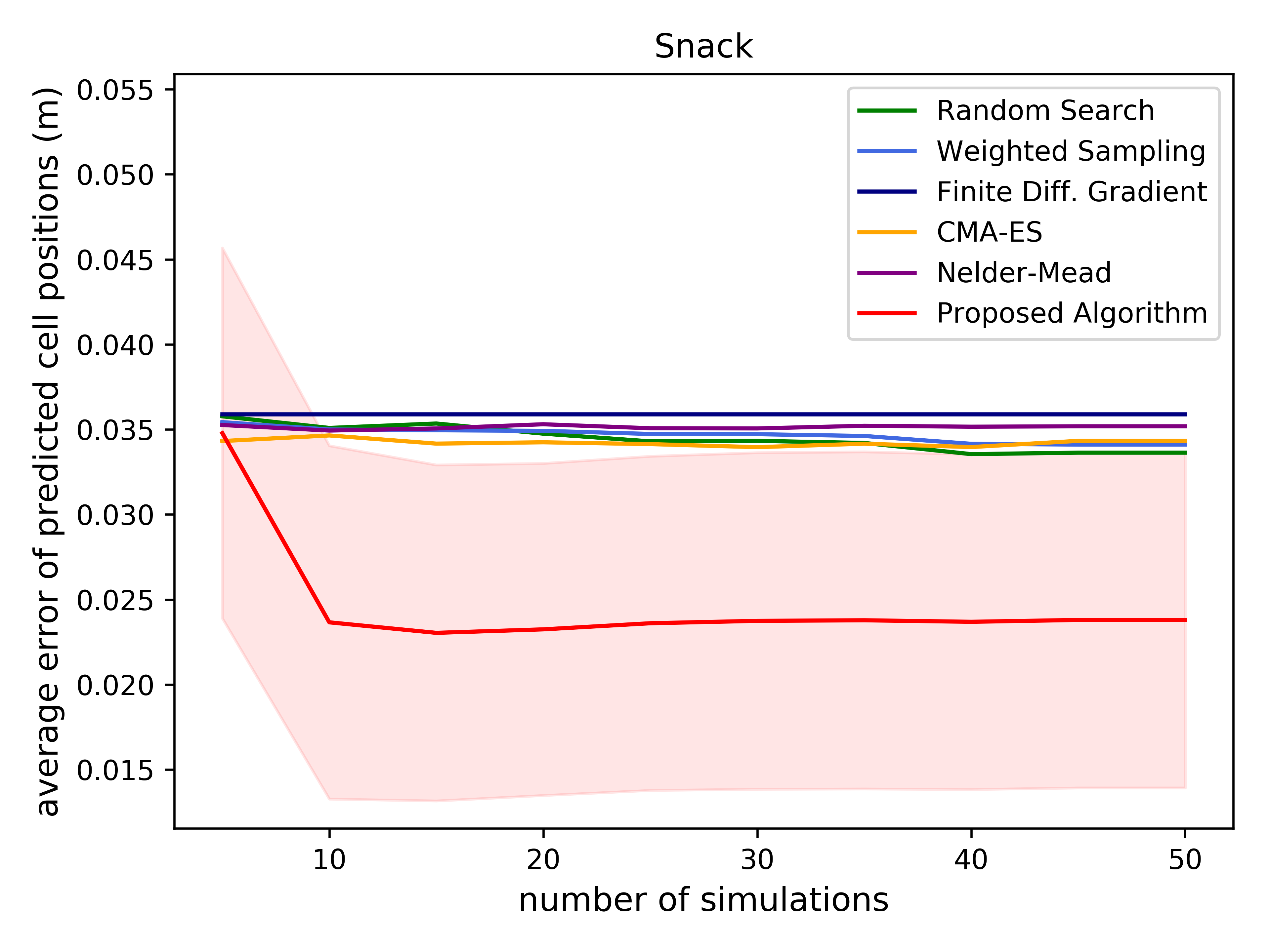} &
 		\includegraphics[width=0.242\textwidth]{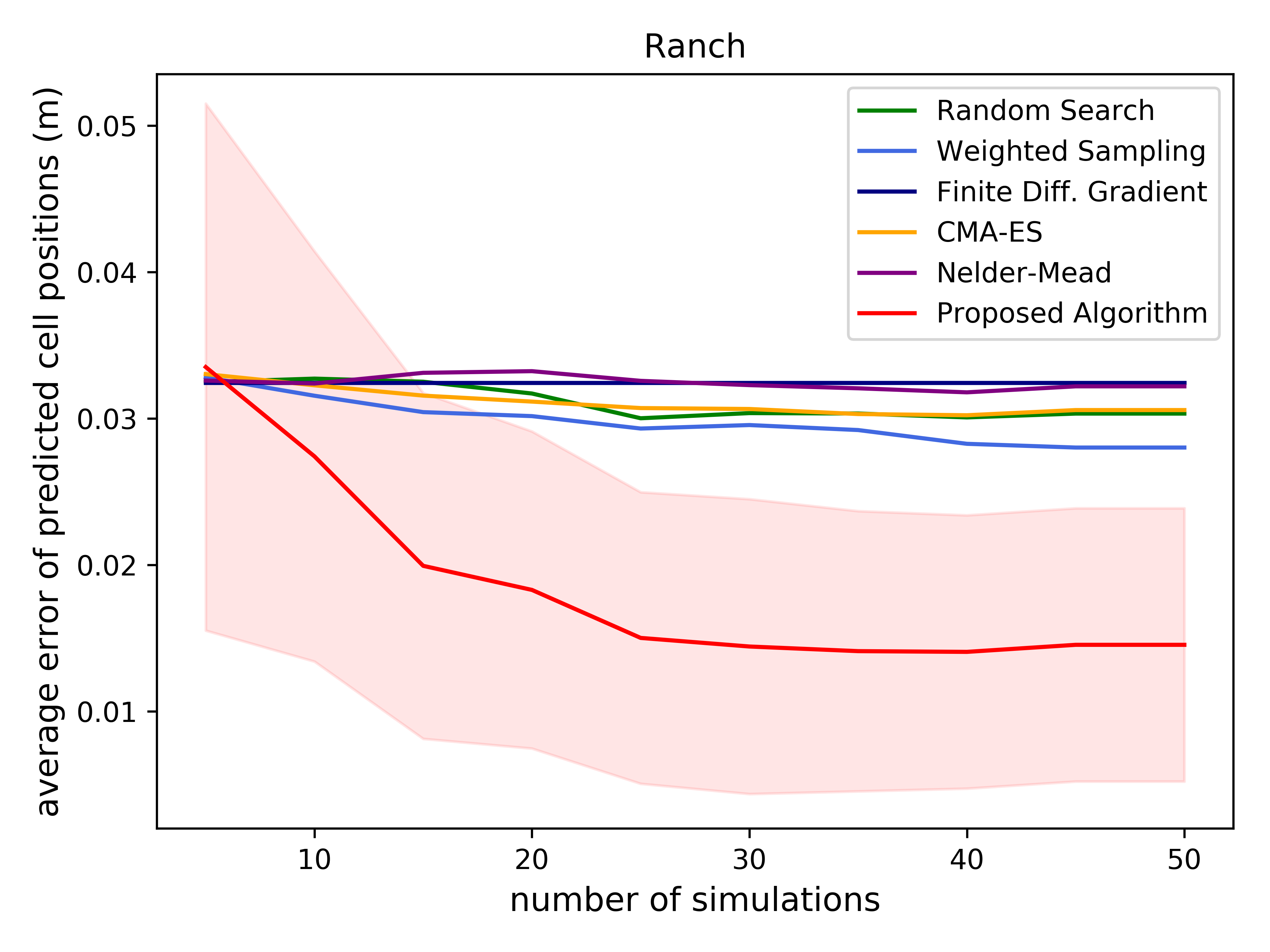} &
 		\includegraphics[width=0.242\textwidth]{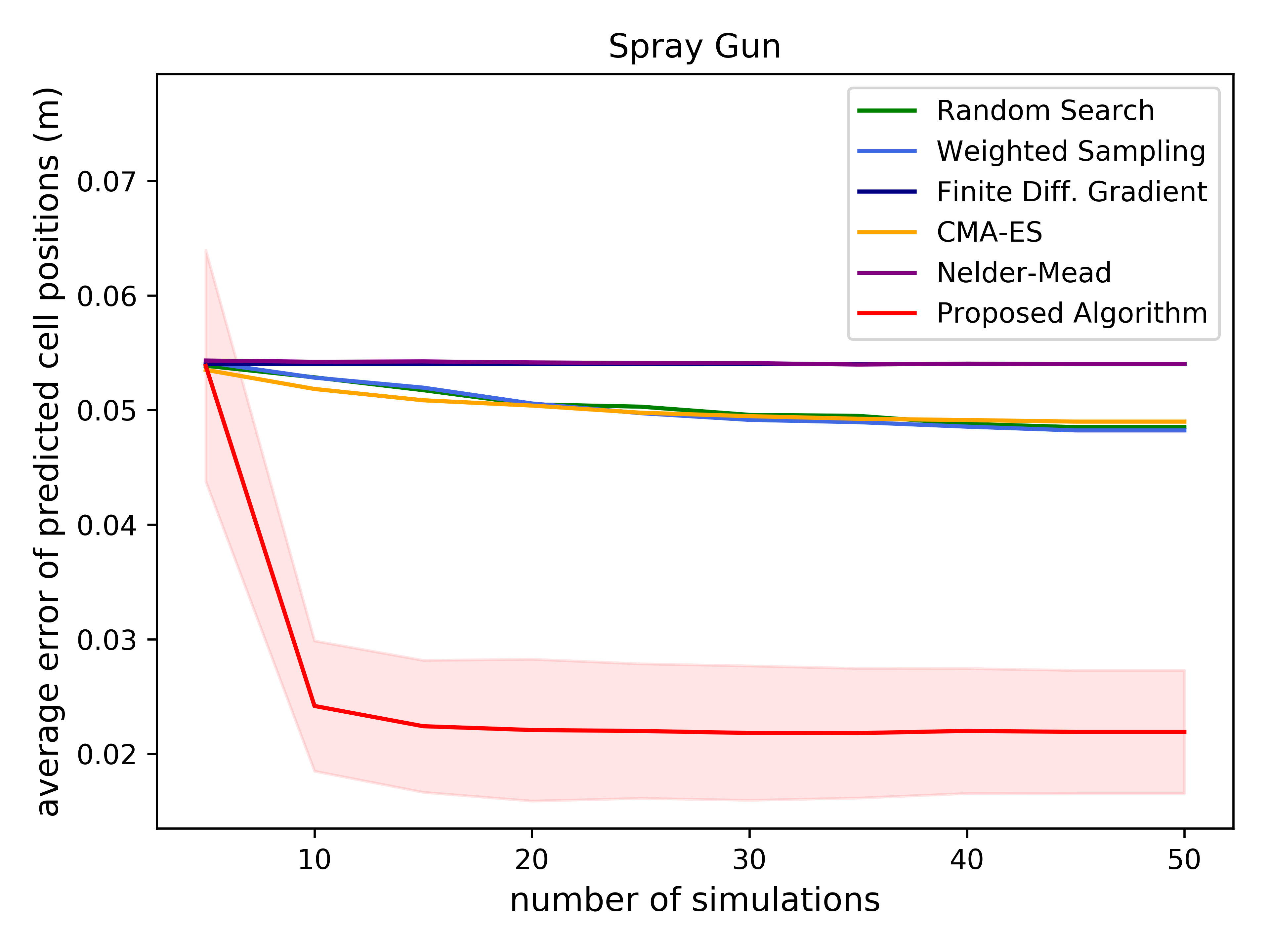} &
 		\includegraphics[width=0.242\textwidth]{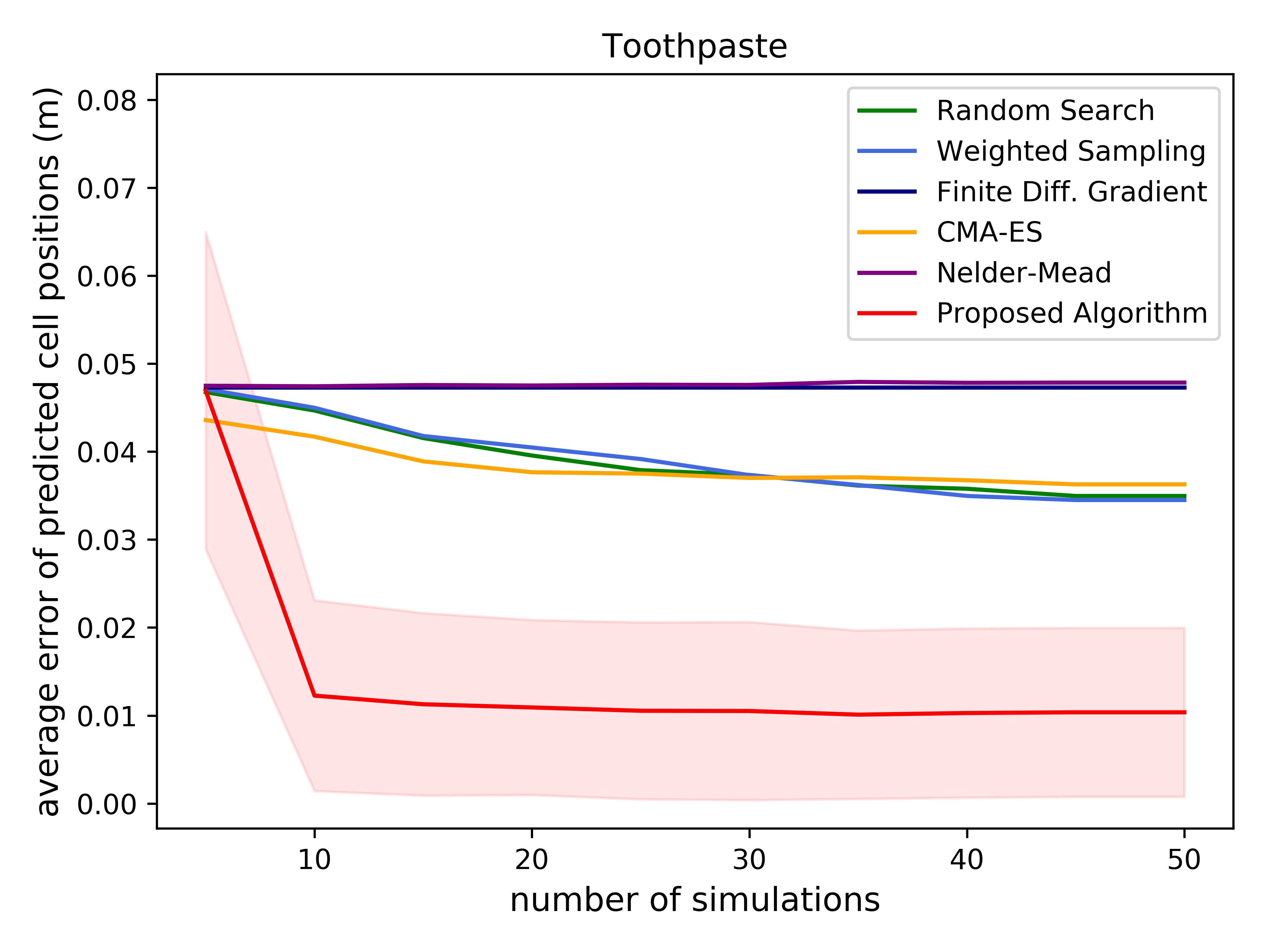}\\
 		(e) & (f) & (g) & (h)\\
 		\includegraphics[width=0.242\textwidth]{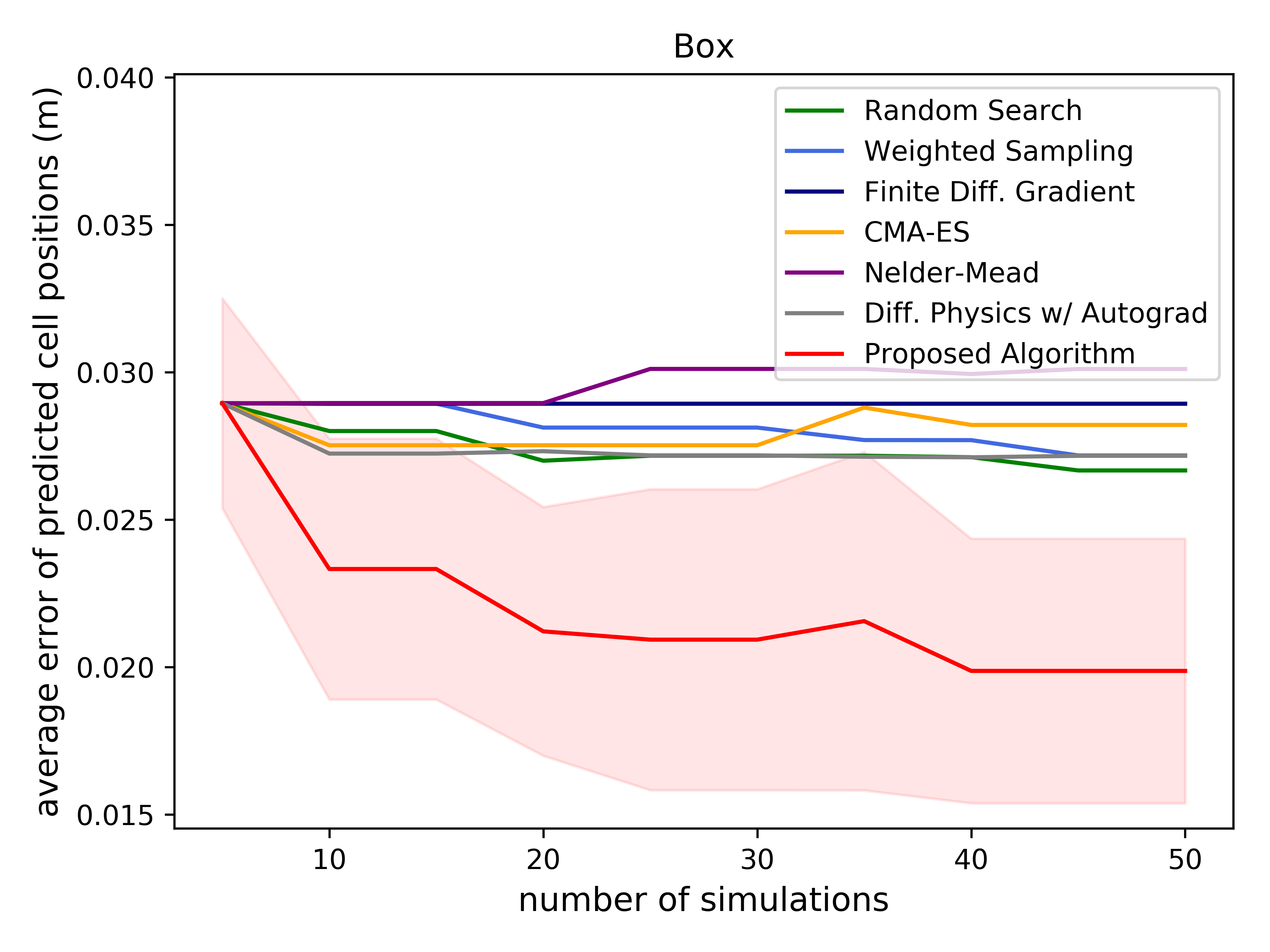} &
 		\includegraphics[width=0.242\textwidth]{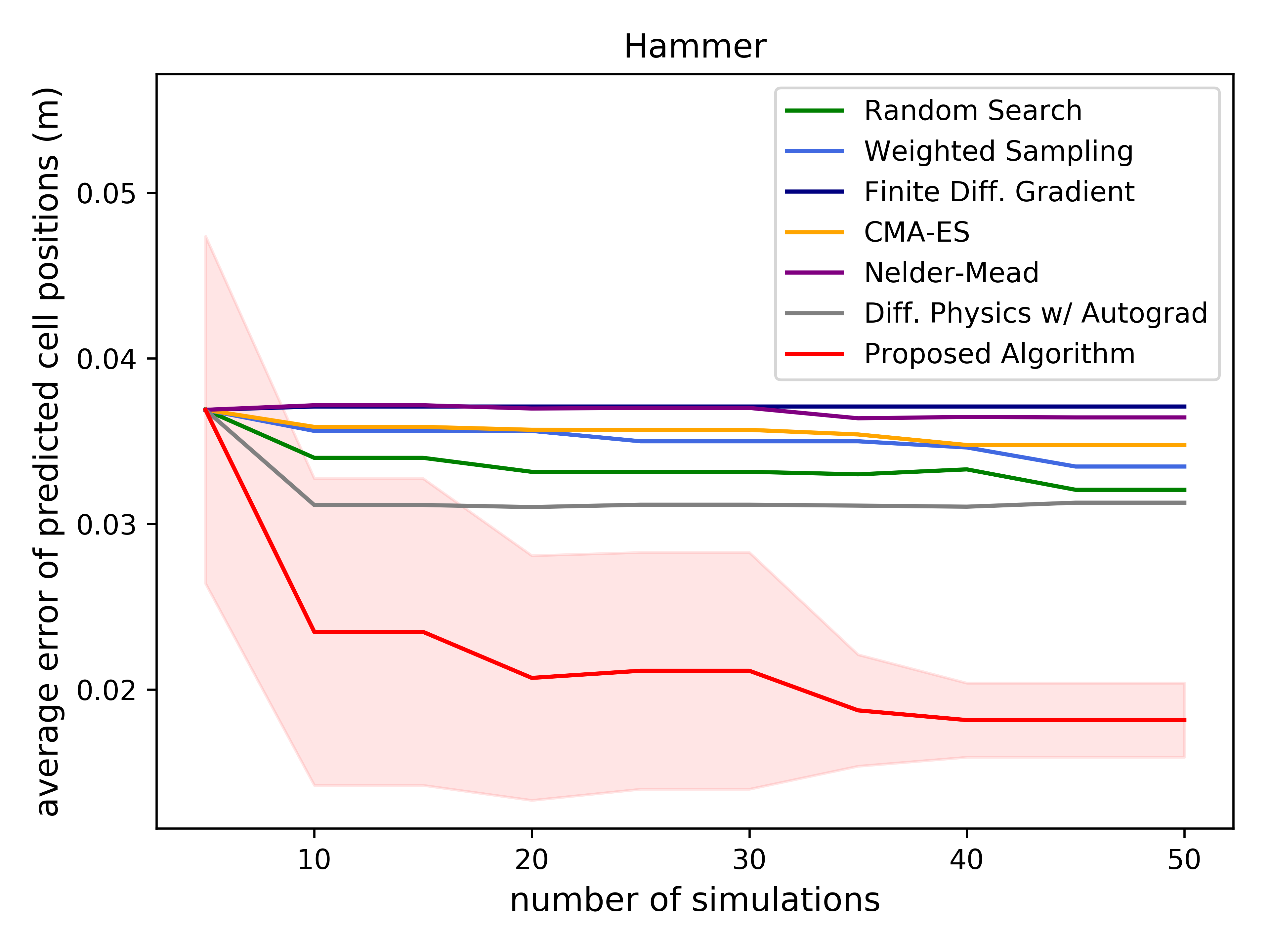} &
		\includegraphics[width=0.242\textwidth]{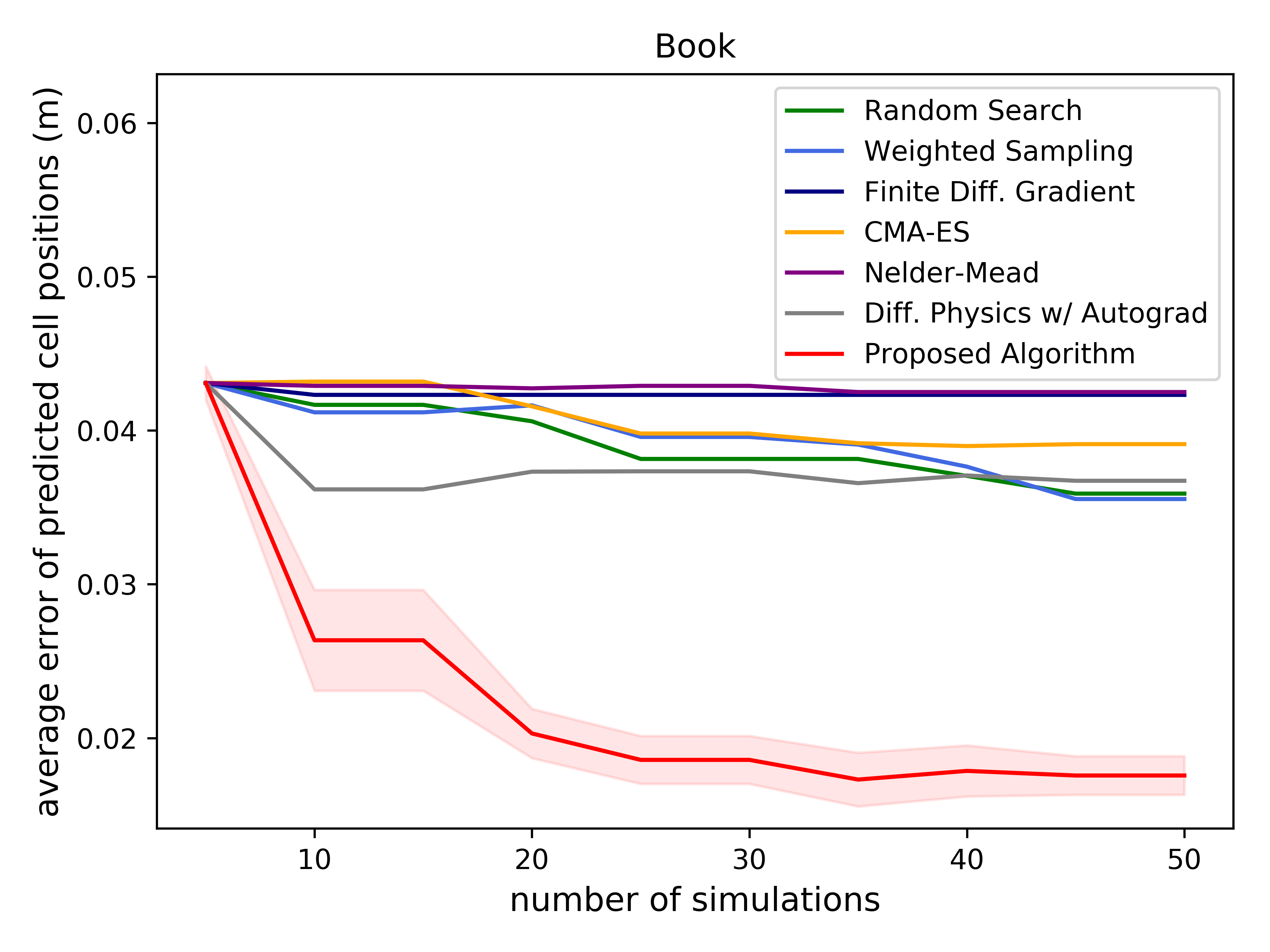} &
 		\includegraphics[width=0.242\textwidth]{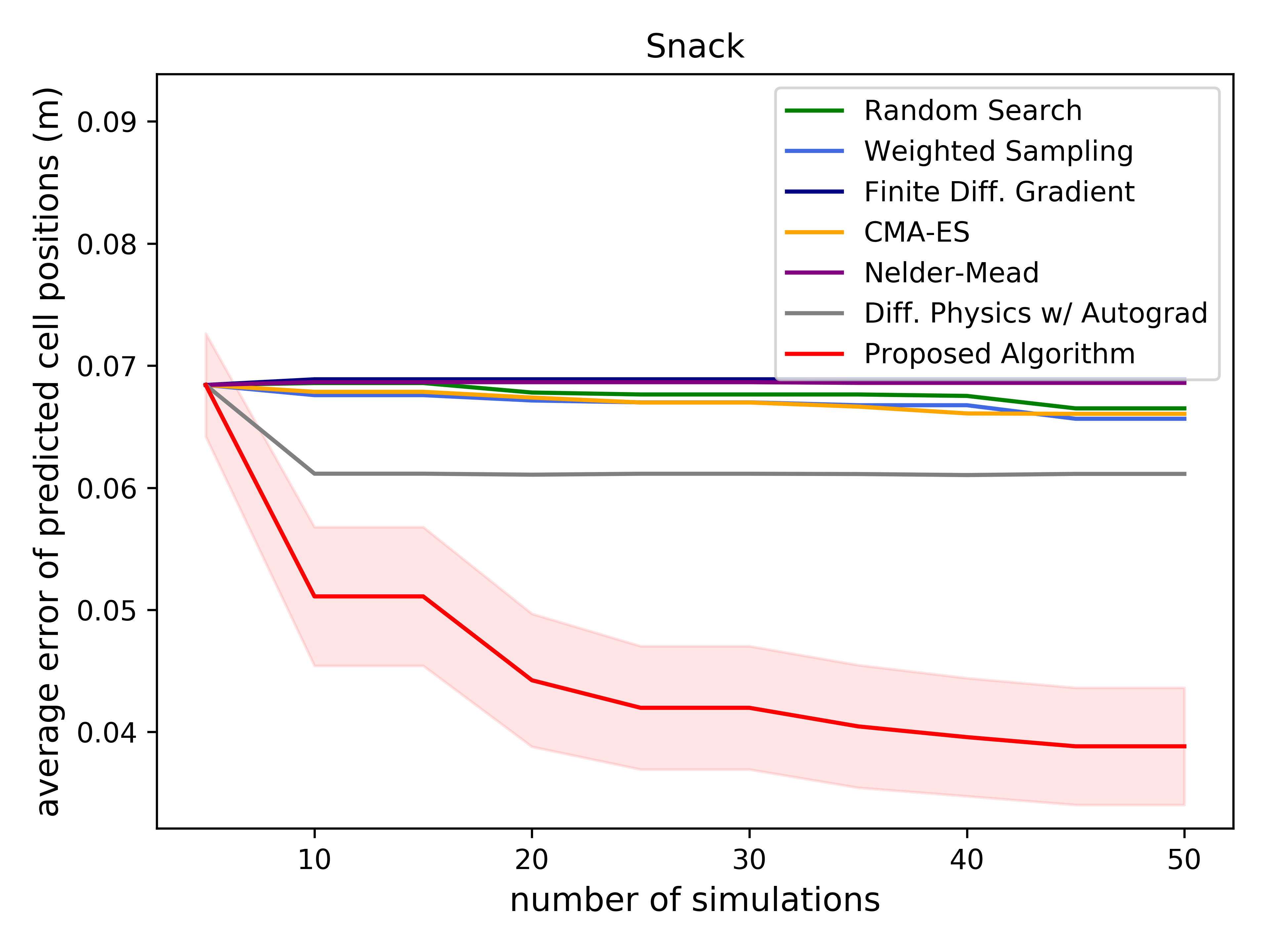} \\
 		(i) & (j) & (k) & (l)\\
		\includegraphics[width=0.242\textwidth]{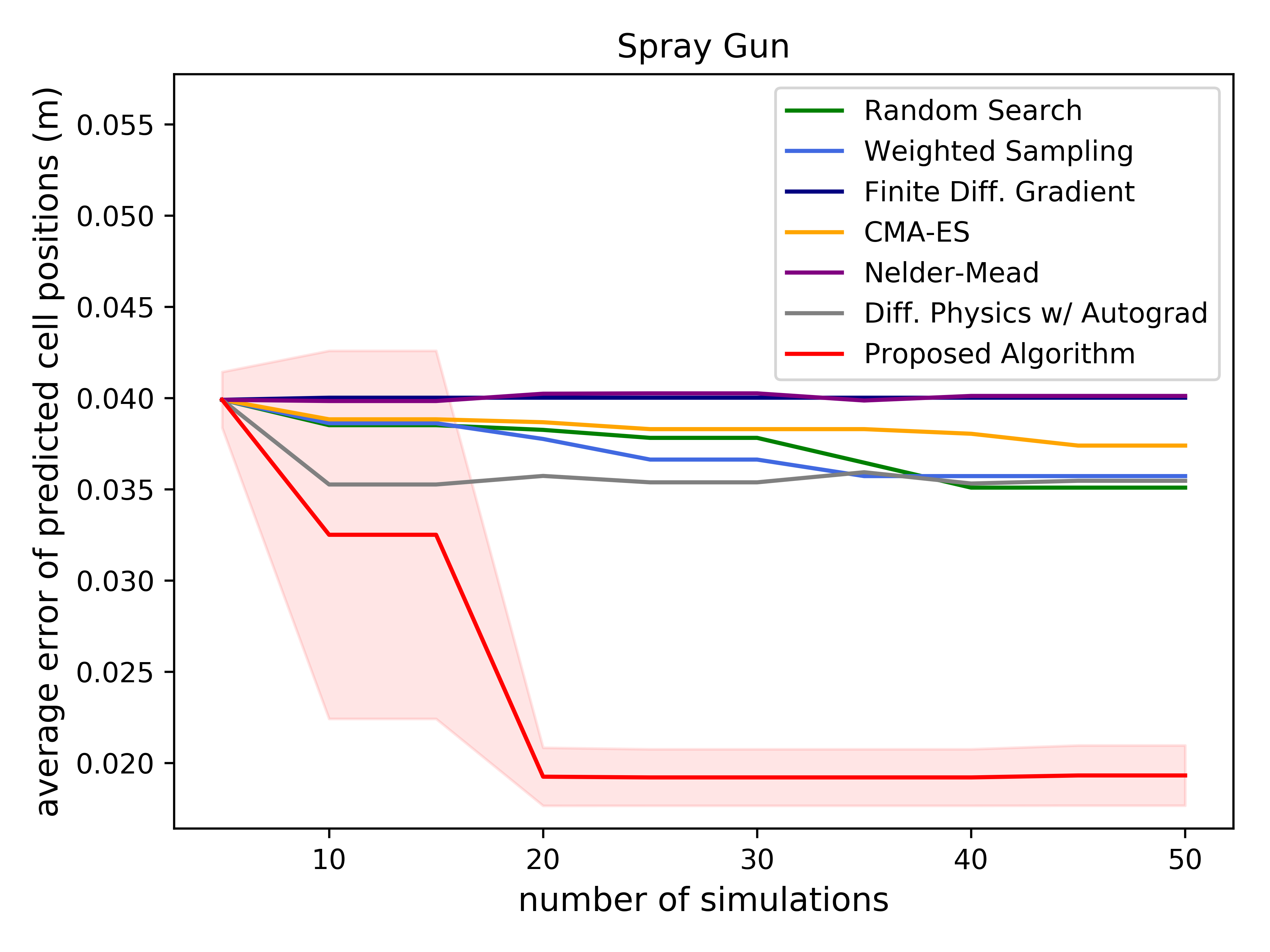} &
		\includegraphics[width=0.242\textwidth]{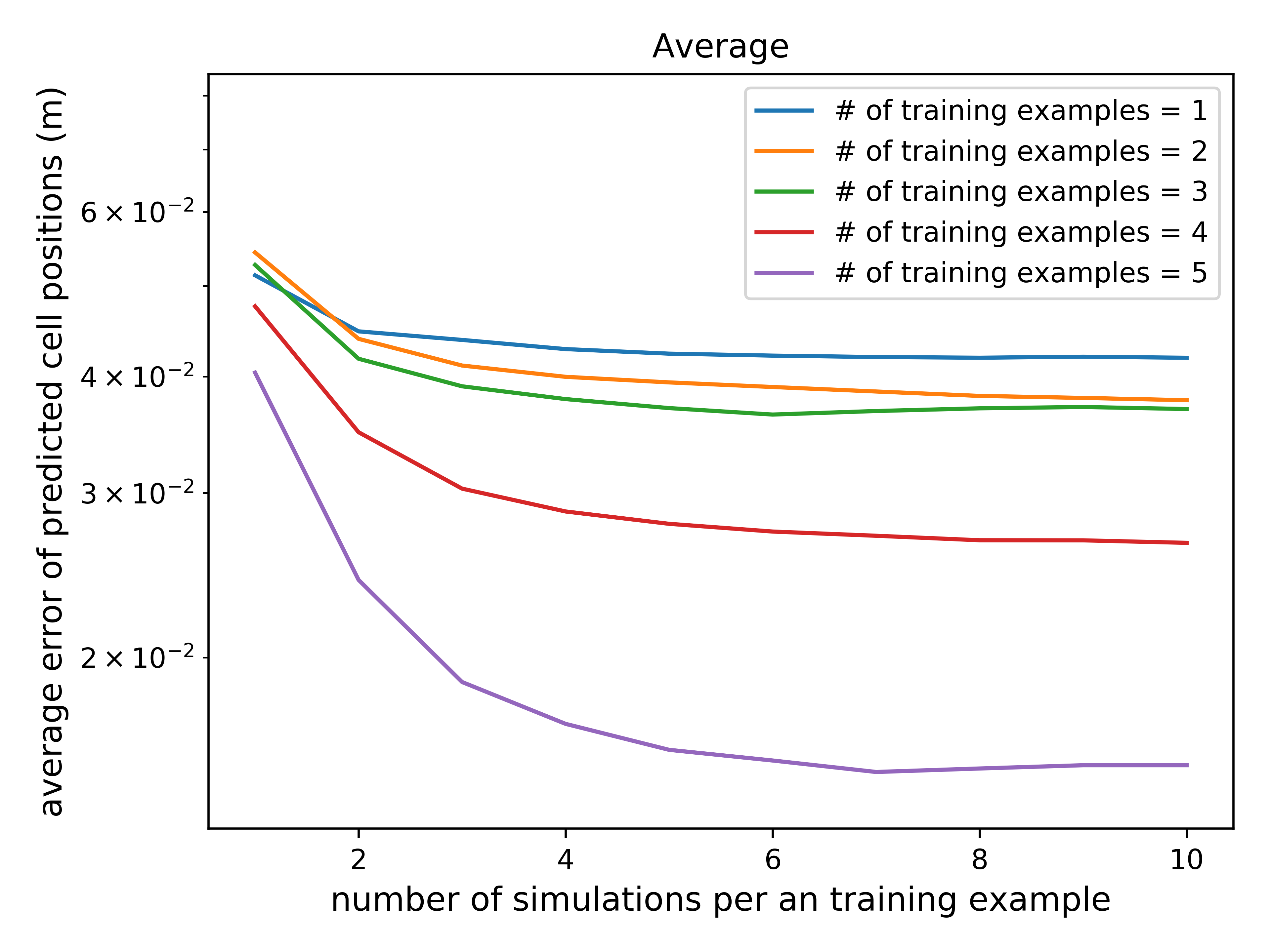} &
		\includegraphics[width=0.242\textwidth]{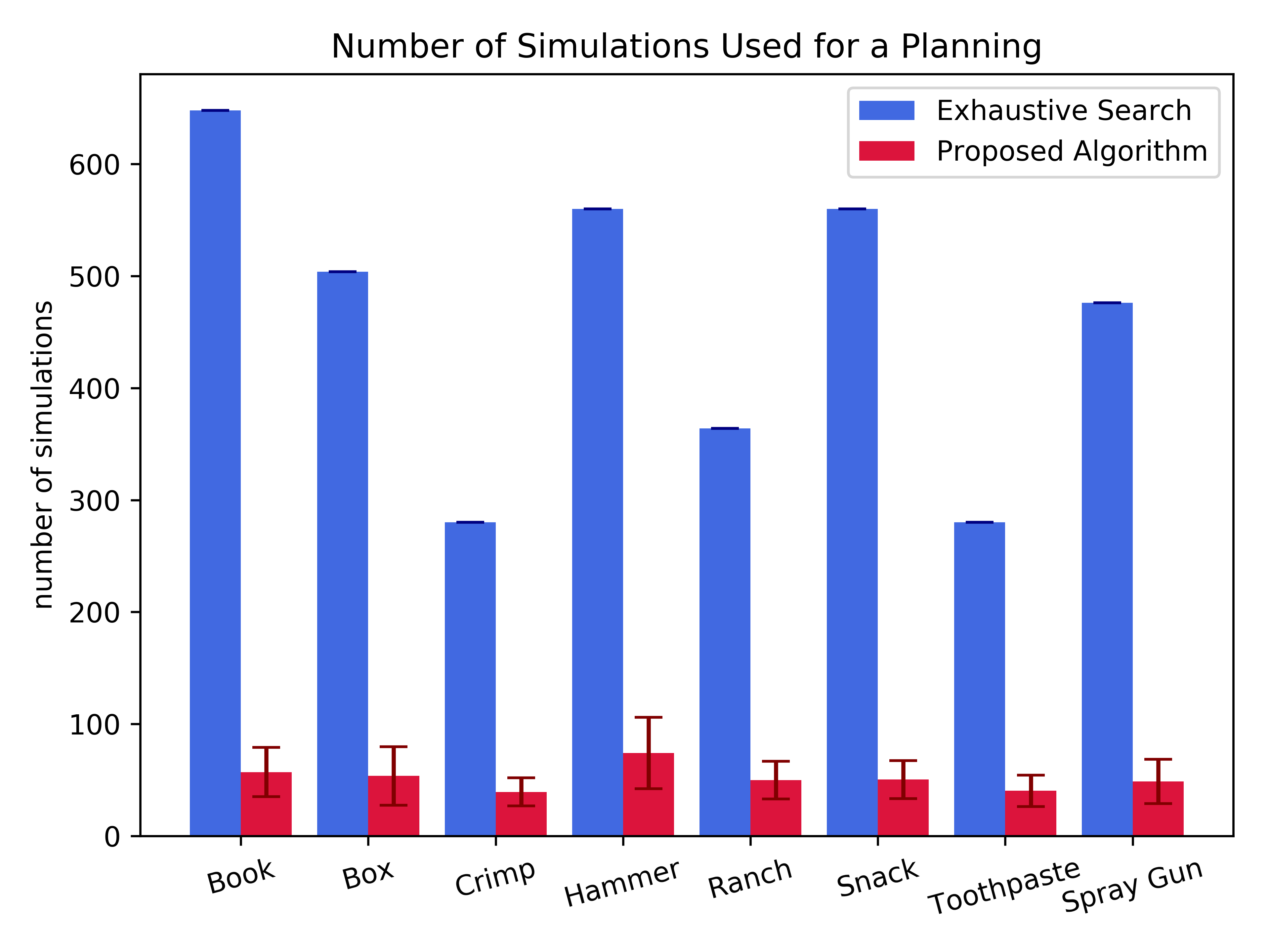} &
		\includegraphics[width=0.242\textwidth]{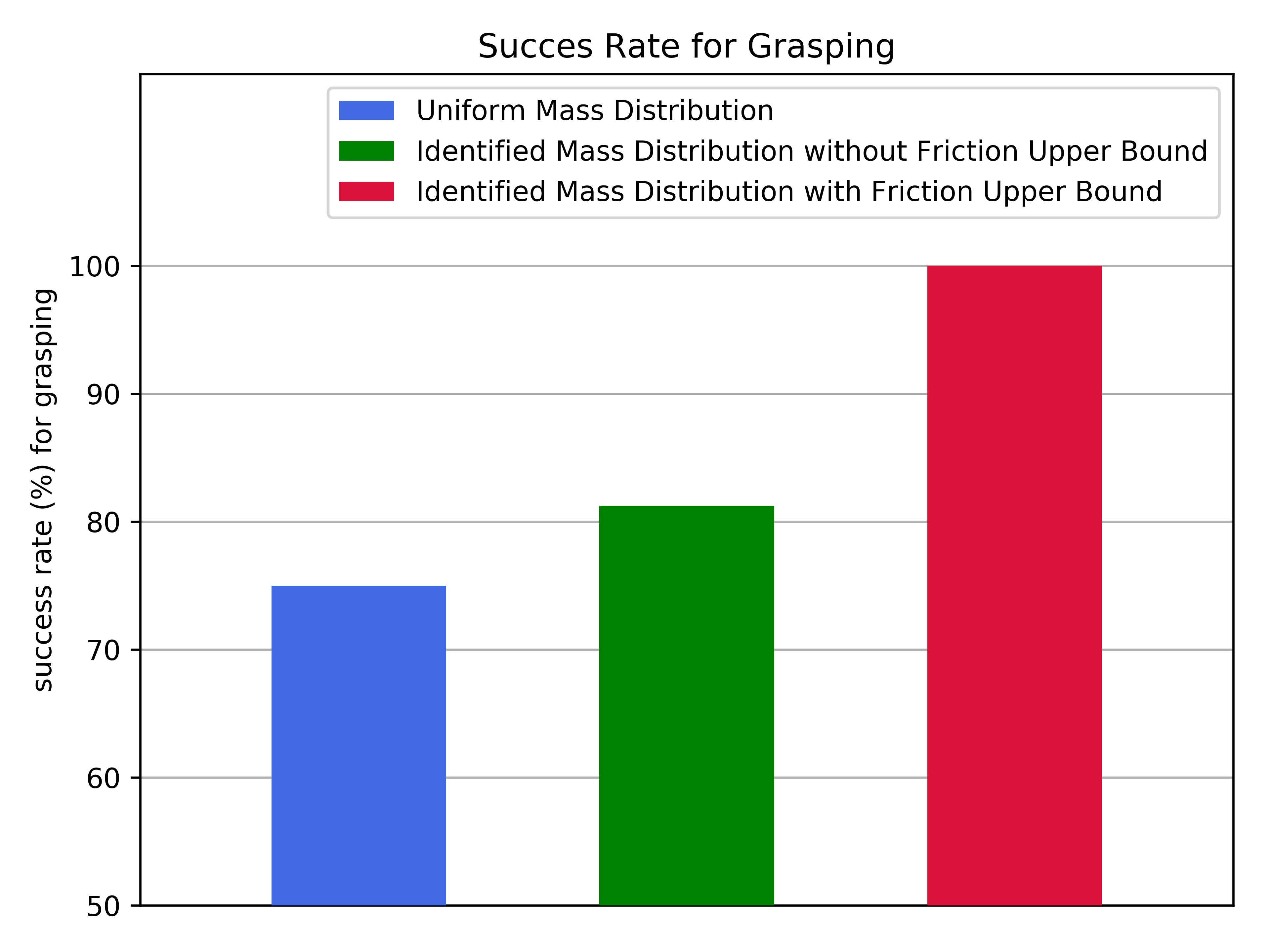} \\
 		(m) & (o) & (p) & (q)\\
	\end{tabular}
    \caption{Average predicted cell position error (in meters) for each object as a function of the number of performed physics simulations. Figures (a)-(h) correspond to the simulation experiments and (i)-(m) correspond to the real experiments. Figure (o) shows the average predicted cell position error with different numbers of training examples. Figure (p) shows the predicted cell position error with different numbers of training examples. Figure (q) shows the success rate for the sliding and grasping task. }
    \vspace{-0.5cm}
    \label{fig:quan_all}
\end{figure*}

\subsection{Results}
{\bf Model Identification}. Figure~\ref{fig:qual_heatmap} shows mass distributions identified using Algorithm~\ref{identificationAlgo}. 
The mass distributions of the book, the hammer, the snack, the toolbox and the spray gun are all identified using the real objects and robot. The experiments on the remaining objects
were performed in simulation because they were too thin for a safe robotic manipulation. The results show that the identified models quickly converge to the ground-truth models. 
Figure~\ref{fig:quan_all}  (a)-(m) shows the average distance between the predicted cell positions and the ground-truth ones in the test data as a function of the number forward physics simulations used by the different identification methods. 
In our method, the number of physics simulations is the same as the number of steps of the gradient-descent, since one simulation is performed after each update of $\mathcal M$ and $\mu$.
Note that the physics simulations dominate the computation time, with $1.3448 (\pm0.5604)$ second per a simulation, while the computation of the gradient using the proposed algorithm takes only $0.0069 (\pm 0.0024)$ second.
The results demonstrate that global optimization methods suffer from the curse of dimensionality due to the combinatorial explosion in the number of possible parameters for all cells. 
The results also demonstrate that the proposed method can estimate the parameters within a surprisingly small number of gradient-descent steps and a short computation time ($30$ seconds).
The average error of the predicted cell positions using the identified mass and friction distributions is less than $1.53cm$ in simulation and $2.27cm$ in the real experiments.
Note that the error in real experiments is higher due to sensing and control errors.
The finite differences approach also failed to converge to an accurate model due to the high computational cost of the gradient computation, as well as the sensitivity of the computed gradients to the choice the grid size. 
Figure \ref{fig:quan_all} (o) shows that the number of training actions improves the accuracy of the model learned by Algorithm~\ref{identificationAlgo}. Increasing the number of training actions allows the robot to uncover properties of different parts of the object more accurately.

{{\bf Planning and Control}.  Figure~\ref{fig:quan_all} (p) shows the number of physics simulations used to optimize the contact location. 
The proposed policy gradient algorithm requires a small fraction of the exhaustive search's computation time, while both methods attained a 100$\%$ success rate in finding a sequence of pushes that reaches the goal in simulation.

{{\bf Pre-grasp Sliding Manipulation}. This task integrates the two previous tasks and evaluates the entire proposed system. Figure~\ref{fig:quan_all} (q) shows that in all of the $16$ trials with random initial poses, 
the robot successfully identified the hammer's mass and friction distributions using Algorithm~\ref{identificationAlgo}, selected a physically stable goal configuration in the desired goal region based on the identified model,  
planned and executed a sequence of pushing actions using Algorithm~\ref{policyAlgo}, and grasped the object from the part pushed out of the table. 
Only $12$ trials resulted in successful grasps when the mass distribution was not learned and was assumed to be uniform. 
The upper bound on the dynamic coefficient of friction is set to $1.00$ in Algorithm~\ref{identificationAlgo}.  
The success rate drops to $13/16$ when no upper bound limit is set on the friction in Algorithm~\ref{identificationAlgo}. 

\section{Final Remark}
The decline in the success rate in Figure~\ref{fig:quan_all} (q)  when no upper limit on the friction is used is an important observation. The algorithm attributed most of the rotations to unrealistically high frictions in certain regions, 
instead of an uneven mass distribution. Despite identifying a wrong model of mass and friction, the predicted motion of the object was accurate as long as the object was entirely on the table. 
But once pushed to the edge, and its heavy side is not anymore supported by the table's surface, the object falls. This clearly demonstrates the importance of identifying not only the friction, 
but also the accurate mass distribution of objects for a safe manipulation.
The upper bound can be seen as an {\it inductive bias} that is necessary for learning from limited data.

\bibliographystyle{unsrtnat}
\bibliography{bibliography}

\begin{thebibliography}{59}
\providecommand{\natexlab}[1]{#1}
\providecommand{\url}[1]{\texttt{#1}}
\expandafter\ifx\csname urlstyle\endcsname\relax
  \providecommand{\doi}[1]{doi: #1}\else
  \providecommand{\doi}{doi: \begingroup \urlstyle{rm}\Url}\fi

\bibitem[Dogar and Srinivasa(2011)]{Dogar2011AFF}
Mehmet~Remzi Dogar and Siddhartha~S. Srinivasa.
\newblock A framework for push-grasping in clutter.
\newblock In \emph{Robotics: Science and Systems}, 2011.

\bibitem[Dogar and Srinivasa(2012)]{Dogar2012}
Mehmet~R. Dogar and Siddhartha~S. Srinivasa.
\newblock A planning framework for non-prehensile manipulation under clutter
  and uncertainty.
\newblock \emph{Autonomous Robots}, 33\penalty0 (3):\penalty0 217--236, Oct
  2012.

\bibitem[King et~al.(2015)King, Haustein, Srinivasa, and Asfour]{king2015}
Jennifer~E. King, Joshua~A. Haustein, Siddhartha~S. Srinivasa, and Tamim
  Asfour.
\newblock Nonprehensile whole arm rearrangement planning on physics manifolds.
\newblock In \emph{{IEEE} International Conference on Robotics and Automation,
  {ICRA} 2015, Seattle, WA, USA, 26-30 May, 2015}, pages 2508--2515. {IEEE},
  2015.
\newblock ISBN 978-1-4799-6923-4.
\newblock \doi{10.1109/ICRA.2015.7139535}.
\newblock URL \url{https://doi.org/10.1109/ICRA.2015.7139535}.

\bibitem[King et~al.(2016)King, Cognetti, and Srinivasa]{king2016}
Jennifer~E. King, Marco Cognetti, and Siddhartha~S. Srinivasa.
\newblock Rearrangement planning using object-centric and robot-centric action
  spaces.
\newblock In \emph{2016 {IEEE} International Conference on Robotics and
  Automation, {ICRA}, Stockholm, Sweden, May 16-21, 2016}, pages 3940--3947,
  2016.

\bibitem[King et~al.(2017)King, Ranganeni, and Srinivasa]{KingICRA2017}
J.~E. King, V.~Ranganeni, and S.~S. Srinivasa.
\newblock Unobservable monte carlo planning for nonprehensile rearrangement
  tasks.
\newblock In \emph{2017 IEEE International Conference on Robotics and
  Automation (ICRA)}, pages 4681--4688, May 2017.
\newblock \doi{10.1109/ICRA.2017.7989544}.

\bibitem[Pinto et~al.(2018)Pinto, Mandalika, Hou, and
  Srinivasa]{Pinto-abs-1810-10654}
Lerrel Pinto, Aditya Mandalika, Brian Hou, and Siddhartha Srinivasa.
\newblock Sample-efficient learning of nonprehensile manipulation policies via
  physics-based informed state distributions.
\newblock \emph{CoRR}, abs/1810.10654, 2018.
\newblock URL \url{http://arxiv.org/abs/1810.10654}.

\bibitem[Yuan et~al.(2018)Yuan, Stork, Kragic, Wang, and
  Hang]{DBLP:conf/icra/YuanSKWH18}
Weihao Yuan, Johannes~A. Stork, Danica Kragic, Michael~Yu Wang, and Kaiyu Hang.
\newblock Rearrangement with nonprehensile manipulation using deep
  reinforcement learning.
\newblock In \emph{2018 {IEEE} International Conference on Robotics and
  Automation, {ICRA} 2018, Brisbane, Australia, May 21-25, 2018}, pages
  270--277, 2018.
\newblock \doi{10.1109/ICRA.2018.8462863}.
\newblock URL \url{https://doi.org/10.1109/ICRA.2018.8462863}.

\bibitem[Mason(1986)]{Mason86}
Matthew~T. Mason.
\newblock Mechanics and planning of manipulator pushing operations.
\newblock \emph{The International Journal of Robotics Research}, 5\penalty0
  (3):\penalty0 53--71, 1986.

\bibitem[N.~Fazeli and Rodriguez(2017)]{fazeli2017ijrr}
R.~Tedrake N.~Fazeli, R.~Kolbert and A.~Rodriguez.
\newblock Parameter and contact force estimation of planar rigid-bodies
  undergoing frictional contact.
\newblock \emph{IJRR}, 2017.

\bibitem[M.~Bauza and Rodriguez(2019)]{bauza2019iros}
Y.~Lin T. Lozano-Perez L. Kaelbling P.~Isola M.~Bauza, F.~Alet and
  A.~Rodriguez.
\newblock Omnipush: accurate, diverse, real-world dataset of pushing dynamics
  with rgbd images.
\newblock In \emph{IROS}, 2019.

\bibitem[Zhou et~al.(2018{\natexlab{a}})Zhou, Mason, Paolini, and
  Bagnell]{JJZhou2018}
Jiaji Zhou, Matthew~T Mason, Robert Paolini, and Drew Bagnell.
\newblock A convex polynomial model for planar sliding mechanics: theory,
  application, and experimental validation.
\newblock \emph{The International Journal of Robotics Research}, 37\penalty0
  (2-3):\penalty0 249--265, 2018{\natexlab{a}}.

\bibitem[Erez et~al.(2015)Erez, Tassa, and Todorov]{ErezTT15}
Tom Erez, Yuval Tassa, and Emanuel Todorov.
\newblock Simulation tools for model-based robotics: Comparison of bullet,
  havok, mujoco, {ODE} and physx.
\newblock In \emph{{IEEE} {ICRA}}, 2015.

\bibitem[DAR()]{DART}
{DART physics egnine}.
\newblock [Online]. Available: \url{ http://dartsim.github.io}.

\bibitem[Phy()]{PhysX}
{PhysX physics engine}.
\newblock [Online]. Available: \url{www.geforce.com/hardware/technology/physx}.

\bibitem[Bul()]{Bullet}
{Bullet physics engine}.
\newblock [Online]. Available: \url{ www.bulletphysics.org}.

\bibitem[ODE()]{ODE}
{Open dynamics engine}.
\newblock [Online]. Available: \url{http://ode.org}.

\bibitem[Dogar et~al.(2012)Dogar, Hsiao, Ciocarlie, and
  Srinivasa]{Dogar_2012_7076}
Mehmet Dogar, Kaijen Hsiao, Matei Ciocarlie, and Siddhartha Srinivasa.
\newblock {Physics-Based Grasp Planning Through Clutter}.
\newblock In \emph{R:SS}, July 2012.

\bibitem[Lynch and Mason(1996{\natexlab{a}})]{LunchMason1996}
K.~M. Lynch and M.~T. Mason.
\newblock Stable pushing: Mechanics, control- lability, and planning.
\newblock \emph{International Journal of Robotics Research}, 18,
  1996{\natexlab{a}}.

\bibitem[Scholz et~al.(2014)Scholz, Levihn, Isbell, and
  Wingate]{isbell:physics:2014}
Jonathan Scholz, Martin Levihn, Charles~L. Isbell, and David Wingate.
\newblock {A Physics-Based Model Prior for Object-Oriented MDPs}.
\newblock In \emph{ICML}, 2014.

\bibitem[Zhou et~al.(2016{\natexlab{a}})Zhou, Paolini, Bagnell, and
  Mason]{ZhouPBM16}
Jiaji Zhou, Robert Paolini, J.~Andrew Bagnell, and Matthew~T. Mason.
\newblock A convex polynomial force-motion model for planar sliding:
  Identification and application.
\newblock In \emph{{ICRA}}, pages 372--377, 2016{\natexlab{a}}.

\bibitem[Abbeel et~al.(2006)Abbeel, Quigley, and Ng]{abbeel2006using}
Pieter Abbeel, Morgan Quigley, and Andrew~Y Ng.
\newblock Using inaccurate models in reinforcement learning.
\newblock In \emph{Proc. of ICML}. ACM, 2006.

\bibitem[Boularias et~al.(2015)Boularias, Bagnell, and
  Stentz]{DBLP:conf/aaai/BoulariasBS15}
Abdeslam Boularias, James~Andrew Bagnell, and Anthony Stentz.
\newblock Learning to manipulate unknown objects in clutter by reinforcement.
\newblock In \emph{Proceedings of the Twenty-Ninth {AAAI} Conference on
  Artificial Intelligence, January 25-30, 2015, Austin, Texas, {USA.}}, pages
  1336--1342, 2015.
\newblock URL
  \url{http://www.aaai.org/ocs/index.php/AAAI/AAAI15/paper/view/9360}.

\bibitem[M.~Bauza* and Rodriguez(2018)]{bauza*2018corl}
F.~Hogan* M.~Bauza* and A.~Rodriguez.
\newblock A data-efficient approach to precise and controlled pushing.
\newblock In \emph{CoRL}, 2018.

\bibitem[Deisenroth et~al.(2011)Deisenroth, Rasmussen, and
  Fox]{Deisenroth:2011fu}
M.~Deisenroth, C.~Rasmussen, and D.~Fox.
\newblock {Learning to Control a Low-Cost Manipulator using Data-Efficient
  Reinforcement Learning}.
\newblock In \emph{R:SS}, 2011.

\bibitem[Calandra et~al.(2016)Calandra, Seyfarth, Peters, and
  Deisenroth]{Calandra2016}
Roberto Calandra, Andr\'e Seyfarth, Jan Peters, and Marc~P. Deisenroth.
\newblock Bayesian optimization for learning gaits under uncertainty.
\newblock \emph{Annals of Mathematics and Artificial Intelligence (AMAI)},
  76\penalty0 (1):\penalty0 5--23, 2016.
\newblock ISSN 1573-7470.

\bibitem[Marco et~al.(2017)Marco, Berkenkamp, Hennig, Schoellig, Krause,
  Schaal, and Trimpe]{MarcoBHS0ST17}
Alonso Marco, Felix Berkenkamp, Philipp Hennig, Angela~P. Schoellig, Andreas
  Krause, Stefan Schaal, and Sebastian Trimpe.
\newblock Virtual vs. real: Trading off simulations and physical experiments in
  reinforcement learning with bayesian optimization.
\newblock In \emph{ICRA}, pages 1557--1563, 2017.

\bibitem[Bansal et~al.(2017)Bansal, Calandra, Xiao, Levine, and
  Tomlin]{bansal2017goal}
Somil Bansal, Roberto Calandra, Ted Xiao, Sergey Levine, and Claire~J Tomlin.
\newblock Goal-driven dynamics learning via bayesian optimization.
\newblock In \emph{IEEE CDC}, 2017.

\bibitem[Pautrat et~al.(2017)Pautrat, Chatzilygeroudis, and
  Mouret]{pautrat2017bayesian}
R{\'e}mi Pautrat, Konstantinos Chatzilygeroudis, and Jean-Baptiste Mouret.
\newblock Bayesian optimization with automatic prior selection for
  data-efficient direct policy search.
\newblock \emph{arXiv preprint arXiv:1709.06919}, 2017.

\bibitem[{Sintov} et~al.(2019){Sintov}, {Morgan}, {Kimmel}, {Dollar}, {Bekris},
  and {Boularias}]{8624443}
A.~{Sintov}, A.~S. {Morgan}, A.~{Kimmel}, A.~M. {Dollar}, K.~E. {Bekris}, and
  A.~{Boularias}.
\newblock Learning a state transition model of an underactuated adaptive hand.
\newblock \emph{IEEE Robotics and Automation Letters}, 4\penalty0 (2):\penalty0
  1287--1294, 2019.

\bibitem[Oh et~al.(2015)Oh, Guo, Lee, Lewis, and Singh]{OhGLLS15}
Junhyuk Oh, Xiaoxiao Guo, Honglak Lee, Richard~L. Lewis, and Satinder~P. Singh.
\newblock Action-conditional video prediction using deep networks in atari
  games.
\newblock In \emph{NIPS}, pages 2863--2871, 2015.

\bibitem[Chiappa et~al.(2017)Chiappa, Racaniere, Wierstra, and
  Mohamed]{chiappa2017recurrent}
Silvia Chiappa, S{\'e}bastien Racaniere, Daan Wierstra, and Shakir Mohamed.
\newblock Recurrent environment simulators.
\newblock \emph{arXiv preprint arXiv:1704.02254}, 2017.

\bibitem[Finn and Levine()]{finn2016deep}
Chelsea Finn and Sergey Levine.
\newblock Deep visual foresight for planning robot motion.
\newblock \emph{ICRA 2017}.

\bibitem[Finn et~al.(2016)Finn, Tan, Duan, Darrell, Levine, and
  Abbeel]{Finn2016DeepSA}
Chelsea Finn, Xin~Yu Tan, Yan Duan, Trevor Darrell, Sergey Levine, and Pieter
  Abbeel.
\newblock Deep spatial autoencoders for visuomotor learning.
\newblock \emph{2016 IEEE International Conference on Robotics and Automation
  (ICRA)}, pages 512--519, 2016.

\bibitem[KT.~Yu and Rodriguez(2016)]{yu2016iros}
N.~Fazeli KT.~Yu, M.~Bauza and A.~Rodriguez.
\newblock More than a million ways to be pushed: A high-fidelity experimental
  dataset of planar pushing.
\newblock In \emph{IROS}, pages 30--37, 2016.

\bibitem[N.~Doshi and Rodriguez(2020)]{doshi2020icra}
F.~Hogan N.~Doshi and A.~Rodriguez.
\newblock Hybrid differential dynamic programming for planar manipulation
  primitives.
\newblock In \emph{ICRA}, 2020.

\bibitem[Lynch and Mason(1996{\natexlab{b}})]{doi:10.1177/027836499601500602}
Kevin~M. Lynch and Matthew~T. Mason.
\newblock Stable pushing: Mechanics, controllability, and planning.
\newblock \emph{The International Journal of Robotics Research}, 15\penalty0
  (6):\penalty0 533--556, 1996{\natexlab{b}}.

\bibitem[Lynch(1993)]{23847993b652419a91558fd1f03bbec3}
Kevin~M. Lynch.
\newblock Estimating the friction parameters of pushed objects.
\newblock In \emph{1993 International Conference on Intelligent Robots and
  Systems}, 1993 International Conference on Intelligent Robots and Systems,
  pages 186--193, 12 1993.

\bibitem[Howe and Cutkosky(1996)]{doi:10.1177/027836499601500603}
Robert~D. Howe and Mark~R. Cutkosky.
\newblock Practical force-motion models for sliding manipulation.
\newblock \emph{The International Journal of Robotics Research}, 15\penalty0
  (6):\penalty0 557--572, 1996.

\bibitem[Dogar and Srinivasa(2010)]{Dogar2010PushgraspingWD}
Mehmet~Remzi Dogar and Siddhartha~S. Srinivasa.
\newblock Push-grasping with dexterous hands: Mechanics and a method.
\newblock \emph{2010 IEEE/RSJ International Conference on Intelligent Robots
  and Systems}, pages 2123--2130, 2010.

\bibitem[Zhou et~al.(2016{\natexlab{b}})Zhou, Paolini, Bagnell, and
  Mason]{DBLP:conf/icra/ZhouPBM16}
Jiaji Zhou, Robert Paolini, J.~Andrew Bagnell, and Matthew~T. Mason.
\newblock A convex polynomial force-motion model for planar sliding:
  Identification and application.
\newblock In \emph{2016 {IEEE} International Conference on Robotics and
  Automation, {ICRA} 2016, Stockholm, Sweden, May 16-21, 2016}, pages 372--377,
  2016{\natexlab{b}}.

\bibitem[Zhou et~al.(2019)Zhou, Hou, and Mason]{DBLP:journals/ijrr/ZhouHM19}
Jiaji Zhou, Yifan Hou, and Matthew~T. Mason.
\newblock Pushing revisited: Differential flatness, trajectory planning, and
  stabilization.
\newblock \emph{I. J. Robotics Res.}, 38\penalty0 (12-13), 2019.

\bibitem[Zhou et~al.(2018{\natexlab{b}})Zhou, Mason, Paolini, and
  Bagnell]{DBLP:journals/ijrr/ZhouMPB18}
Jiaji Zhou, Matthew~T. Mason, Robert Paolini, and Drew Bagnell.
\newblock A convex polynomial model for planar sliding mechanics: theory,
  application, and experimental validation.
\newblock \emph{I. J. Robotics Res.}, 37\penalty0 (2-3):\penalty0 249--265,
  2018{\natexlab{b}}.

\bibitem[Yoshikawa and Kurisu(1991)]{Yoshikawa1991IndentificationOT}
Tsuneo Yoshikawa and Masamitsu Kurisu.
\newblock Indentification of the center of friction from pushing an object by a
  mobile robot.
\newblock \emph{Proceedings IROS '91:IEEE/RSJ International Workshop on
  Intelligent Robots and Systems '91}, pages 449--454 vol.2, 1991.

\bibitem[Zhu et~al.(2018)Zhu, Kimmel, Bekris, and Boularias]{ShaojunIJCAI2018}
Shaojun Zhu, Andrew Kimmel, Kostas Bekris, and Abdeslam Boularias.
\newblock Fast model identification via physics engines for improved policy
  search.
\newblock In \emph{Proceedings of the 27th International Joint Conference on
  Artificial Intelligence (IJCAI), Stockholm, Sweden}, 2018.

\bibitem[Song and Boularias(2020)]{L4DC2020Changkyu}
Changkyu Song and Abdeslam Boularias.
\newblock Identifying mechanical models of unknown objects with differentiable
  physics simulations.
\newblock In \emph{Proceedings of the 2020 Learning for Dynamics and Control
  Conference (L4DC), Berkeley, California, 2020}, 2020.

\bibitem[Degrave et~al.(2016)Degrave, Hermans, Dambre, and
  Wyffels]{DegraveHDW16}
Jonas Degrave, Michiel Hermans, Joni Dambre, and Francis Wyffels.
\newblock A differentiable physics engine for deep learning in robotics.
\newblock \emph{CoRR}, abs/1611.01652, 2016.

\bibitem[Al{-}Rfou et~al.(2016)Al{-}Rfou, Alain, Almahairi, Angerm{\"{u}}ller,
  Bahdanau, Ballas, Bastien, Bayer, Belikov, Belopolsky, Bengio, Bergeron,
  Bergstra, Bisson, Snyder, Bouchard, Boulanger{-}Lewandowski, Bouthillier,
  de~Br{\'{e}}bisson, Breuleux, Carrier, Cho, Chorowski, Christiano, Cooijmans,
  C{\^{o}}t{\'{e}}, C{\^{o}}t{\'{e}}, Courville, Dauphin, Delalleau, Demouth,
  Desjardins, Dieleman, Dinh, Ducoffe, Dumoulin, Kahou, Erhan, Fan, Firat,
  Germain, Glorot, Goodfellow, Graham, G{\"{u}}l{\c{c}}ehre, Hamel, Harlouchet,
  Heng, Hidasi, Honari, Jain, Jean, Jia, Korobov, Kulkarni, Lamb, Lamblin,
  Larsen, Laurent, Lee, Lefran{\c{c}}ois, Lemieux, L{\'{e}}onard, Lin, Livezey,
  Lorenz, Lowin, Ma, Manzagol, Mastropietro, McGibbon, Memisevic, van
  Merri{\"{e}}nboer, Michalski, Mirza, Orlandi, Pal, Pascanu, Pezeshki, Raffel,
  Renshaw, Rocklin, Romero, Roth, Sadowski, Salvatier, Savard, Schl{\"{u}}ter,
  Schulman, Schwartz, Serban, Serdyuk, Shabanian, Simon, Spieckermann,
  Subramanyam, Sygnowski, Tanguay, van Tulder, Turian, Urban, Vincent, Visin,
  de~Vries, Warde{-}Farley, Webb, Willson, Xu, Xue, Yao, Zhang, and
  Zhang]{DBLP:journals/corr/Al-RfouAAa16}
Rami Al{-}Rfou, Guillaume Alain, Amjad Almahairi, Christof Angerm{\"{u}}ller,
  Dzmitry Bahdanau, Nicolas Ballas, Fr{\'{e}}d{\'{e}}ric Bastien, Justin Bayer,
  Anatoly Belikov, Alexander Belopolsky, Yoshua Bengio, Arnaud Bergeron, James
  Bergstra, Valentin Bisson, Josh~Bleecher Snyder, Nicolas Bouchard, Nicolas
  Boulanger{-}Lewandowski, Xavier Bouthillier, Alexandre de~Br{\'{e}}bisson,
  Olivier Breuleux, Pierre~Luc Carrier, Kyunghyun Cho, Jan Chorowski, Paul~F.
  Christiano, Tim Cooijmans, Marc{-}Alexandre C{\^{o}}t{\'{e}}, Myriam
  C{\^{o}}t{\'{e}}, Aaron~C. Courville, Yann~N. Dauphin, Olivier Delalleau,
  Julien Demouth, Guillaume Desjardins, Sander Dieleman, Laurent Dinh, Melanie
  Ducoffe, Vincent Dumoulin, Samira~Ebrahimi Kahou, Dumitru Erhan, Ziye Fan,
  Orhan Firat, Mathieu Germain, Xavier Glorot, Ian~J. Goodfellow, Matthew
  Graham, {\c{C}}aglar G{\"{u}}l{\c{c}}ehre, Philippe Hamel, Iban Harlouchet,
  Jean{-}Philippe Heng, Bal{\'{a}}zs Hidasi, Sina Honari, Arjun Jain,
  S{\'{e}}bastien Jean, Kai Jia, Mikhail Korobov, Vivek Kulkarni, Alex Lamb,
  Pascal Lamblin, Eric Larsen, C{\'{e}}sar Laurent, Sean Lee, Simon
  Lefran{\c{c}}ois, Simon Lemieux, Nicholas L{\'{e}}onard, Zhouhan Lin,
  Jesse~A. Livezey, Cory Lorenz, Jeremiah Lowin, Qianli Ma, Pierre{-}Antoine
  Manzagol, Olivier Mastropietro, Robert McGibbon, Roland Memisevic, Bart van
  Merri{\"{e}}nboer, Vincent Michalski, Mehdi Mirza, Alberto Orlandi,
  Christopher~Joseph Pal, Razvan Pascanu, Mohammad Pezeshki, Colin Raffel,
  Daniel Renshaw, Matthew Rocklin, Adriana Romero, Markus Roth, Peter Sadowski,
  John Salvatier, Fran{\c{c}}ois Savard, Jan Schl{\"{u}}ter, John Schulman,
  Gabriel Schwartz, Iulian~Vlad Serban, Dmitriy Serdyuk, Samira Shabanian,
  {\'{E}}tienne Simon, Sigurd Spieckermann, S.~Ramana Subramanyam, Jakub
  Sygnowski, J{\'{e}}r{\'{e}}mie Tanguay, Gijs van Tulder, Joseph~P. Turian,
  Sebastian Urban, Pascal Vincent, Francesco Visin, Harm de~Vries, David
  Warde{-}Farley, Dustin~J. Webb, Matthew Willson, Kelvin Xu, Lijun Xue,
  Li~Yao, Saizheng Zhang, and Ying Zhang.
\newblock Theano: {A} python framework for fast computation of mathematical
  expressions.
\newblock \emph{CoRR}, abs/1605.02688, 2016.

\bibitem[Kloss et~al.(2017)Kloss, Schaal, and
  Bohg]{DBLP:journals/corr/abs-1710-04102}
Alina Kloss, Stefan Schaal, and Jeannette Bohg.
\newblock Combining learned and analytical models for predicting action
  effects.
\newblock \emph{CoRR}, abs/1710.04102, 2017.

\bibitem[Schenck and Fox(2018)]{Schenck2018SPNetsDF}
Connor Schenck and Dieter Fox.
\newblock Spnets: Differentiable fluid dynamics for deep neural networks.
\newblock \emph{CoRR}, abs/1806.06094, 2018.

\bibitem[Toussaint et~al.(2018)Toussaint, Allen, Smith, and
  Tenenbaum]{18-toussaint-RSS}
Marc Toussaint, Kelsey~R Allen, Kevin~A Smith, and Josh~B Tenenbaum.
\newblock Differentiable physics and stable modes for tool-use and manipulation
  planning.
\newblock In \emph{Proc{.} of Robotics: Science and Systems (R:SS 2018)}, 2018.

\bibitem[de~Avila Belbute-Peres and Kolter(2017)]{Belbute-Peres2017}
Filipe de~Avila Belbute-Peres and Zico Kolter.
\newblock A modular differentiable rigid body physics engine.
\newblock In \emph{Deep Reinforcement Learning Symposium, NIPS}, 2017.

\bibitem[Mordatch et~al.(2012)Mordatch, Todorov, and
  Popovi\'{c}]{Mordatch:2012}
Igor Mordatch, Emanuel Todorov, and Zoran Popovi\'{c}.
\newblock Discovery of complex behaviors through contact-invariant
  optimization.
\newblock \emph{ACM Trans. Graph.}, 31\penalty0 (4):\penalty0 43:1--43:8, July
  2012.
\newblock ISSN 0730-0301.
\newblock \doi{10.1145/2185520.2185539}.
\newblock URL \url{http://doi.acm.org/10.1145/2185520.2185539}.

\bibitem[Hinterstoisser et~al.(2013)Hinterstoisser, Lepetit, Ilic, Holzer,
  Bradski, Konolige, and Navab]{Hinterstoisser2013}
Stefan Hinterstoisser, Vincent Lepetit, Slobodan Ilic, Stefan Holzer, Gary
  Bradski, Kurt Konolige, and Nassir Navab.
\newblock Model based training, detection and pose estimation of texture-less
  3d objects in heavily cluttered scenes.
\newblock In Kyoung~Mu Lee, Yasuyuki Matsushita, James~M. Rehg, and Zhanyi Hu,
  editors, \emph{Computer Vision -- ACCV 2012}, pages 548--562, 2013.

\bibitem[Cline(2002)]{cline}
Michael~Bradley Cline.
\newblock \emph{Rigid body simulation with contact and constraints}.
\newblock PhD thesis, University of British Columbia, 2002.

\bibitem[Belbute-Peres et~al.(2018)Belbute-Peres, Smith, Allen, Tenenbaum, and
  Kolter]{Belbute-Peres:2018:EDP:3327757.3327820}
Filipe de~A. Belbute-Peres, Kevin~A. Smith, Kelsey~R. Allen, Joshua~B.
  Tenenbaum, and J.~Zico Kolter.
\newblock End-to-end differentiable physics for learning and control.
\newblock In \emph{Proceedings of the 32Nd International Conference on Neural
  Information Processing Systems}, NIPS'18, pages 7178--7189, USA, 2018. Curran
  Associates Inc.
\newblock URL \url{http://dl.acm.org/citation.cfm?id=3327757.3327820}.

\bibitem[Mattingley and Boyd(2012)]{MattingleySB12}
J.~Mattingley and S.~Boyd.
\newblock {CVXGEN}: A code generator for embedded convex optimization.
\newblock \emph{Optimization and Engineering}, 12\penalty0 (1):\penalty0 1--27,
  2012.

\bibitem[Hang et~al.(2019)Hang, Morgan, and Dollar]{Kaiyu2019}
Kaiyu Hang, Andrew Morgan, and Aaron Dollar.
\newblock Pre-grasp sliding manipulation of thin objects using soft, compliant,
  or underactuated hands.
\newblock \emph{IEEE Robotics and Automation Letters}, PP:\penalty0 1--1, 01
  2019.
\newblock \doi{10.1109/LRA.2019.2892591}.

\bibitem[Hansen(2006)]{hansen2006cma}
Nikolaus Hansen.
\newblock The cma evolution strategy: a comparing review.
\newblock \emph{Towards a new evolutionary computation}, pages 75--102, 2006.

\bibitem[Paszke et~al.(2017)Paszke, Gross, Chintala, Chanan, Yang, DeVito, Lin,
  Desmaison, Antiga, and Lerer]{paszke2017automatic}
Adam Paszke, Sam Gross, Soumith Chintala, Gregory Chanan, Edward Yang, Zachary
  DeVito, Zeming Lin, Alban Desmaison, Luca Antiga, and Adam Lerer.
\newblock Automatic differentiation in pytorch.
\newblock 2017.

\end{thebibliography}
\end{document}